\documentclass[10pt,twocolumn,letterpaper]{article}

\usepackage{arxiv}
\usepackage{times}
\usepackage{epsfig}
\usepackage{graphicx}
\usepackage{amsmath}
\usepackage{amssymb}
\usepackage{subcaption}

\usepackage[pagebackref=true,breaklinks=true,letterpaper=true,colorlinks,bookmarks=false]{hyperref}

\begin{document}

\title{Recovering Geometric Information with Learned Texture Perturbations}
\author{
Jane Wu\textsuperscript{1} \qquad Yongxu Jin\textsuperscript{1} \qquad Zhenglin Geng\textsuperscript{1} \qquad Hui Zhou\textsuperscript{2,$\dagger$} \qquad Ronald Fedkiw\textsuperscript{1,3} \\
\textsuperscript{1}Stanford University \qquad \textsuperscript{2}JD.com \qquad \textsuperscript{3}Industrial Light \& Magic \\
{\tt\small \textsuperscript{1}\{janehwu,yxjin,zhenglin,rfedkiw\}@stanford.edu \qquad \textsuperscript{$\dagger$}hui.zhou@jd.com}
}

\maketitle

\begin{abstract}
Regularization is used to avoid overfitting when training a neural network; unfortunately, this reduces the attainable level of detail hindering the ability to capture high-frequency information present in the training data.
Even though various approaches may be used to re-introduce high-frequency detail, it typically does not match the training data and is often not time coherent.
In the case of network inferred cloth, these sentiments manifest themselves via either a lack of detailed wrinkles or unnaturally appearing and/or time incoherent surrogate wrinkles.
Thus, we propose a general strategy whereby high-frequency information is procedurally embedded into low-frequency data so that when the latter is smeared out by the network the former still retains its high-frequency detail.
We illustrate this approach by learning texture coordinates which when smeared do not in turn smear out the high-frequency detail in the texture itself but merely smoothly distort it.
Notably, we prescribe perturbed texture coordinates that are subsequently used to correct the over-smoothed appearance of inferred cloth, and correcting the appearance from multiple camera views naturally recovers lost geometric information.
\end{abstract}

\section{Introduction}
Since neural networks are trained to generalize to unseen data, regularization is important for reducing overfitting, see \eg \cite{goodfellow2016deep,scholkopf2001learning}.
However, regularization also removes some of the high variance characteristic of much of the physical world.
Even though high-quality ground truth data can be collected or generated to reflect the desired complexity of the outputs, regularization will inevitably smooth network predictions.
Rather than attempting to directly infer high-frequency features, we alternatively propose to learn a low-frequency space in which such features can be embedded.

\begin{figure}[t]
\centering
    \begin{subfigure}[b]{0.48\linewidth}
        \includegraphics[width=\linewidth]{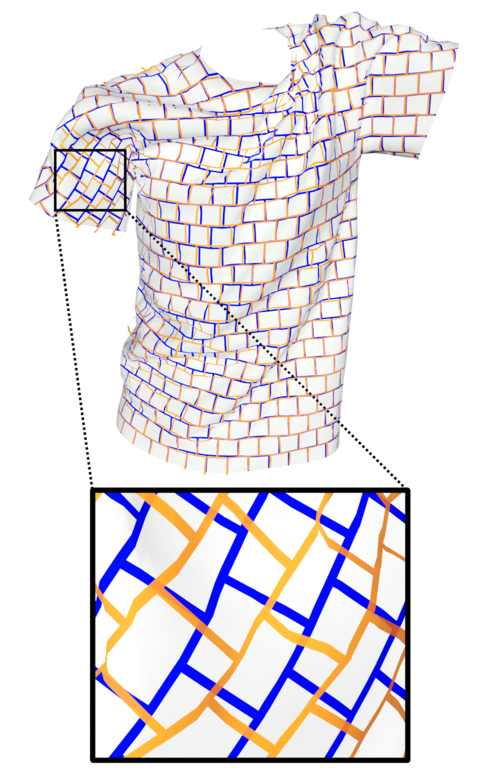}
        \caption{inferred cloth}
        \label{fig:sec1_pd}
    \end{subfigure}
    \begin{subfigure}[b]{0.48\linewidth}
        \includegraphics[width=\linewidth]{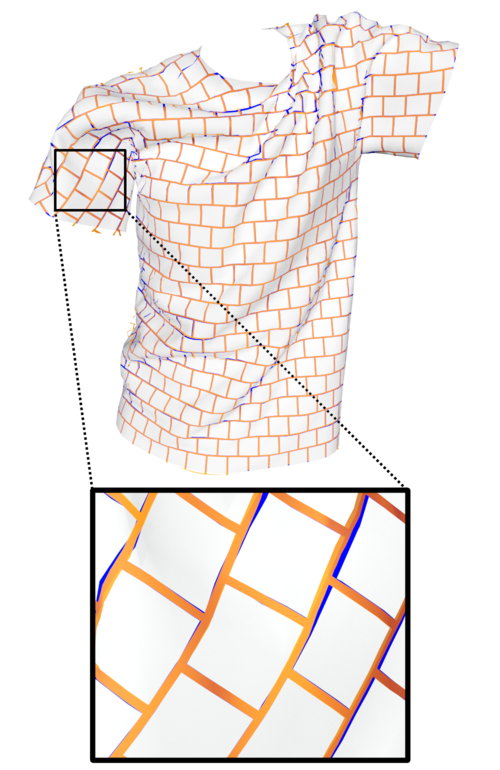}
        \caption{texture sliding}
        \label{fig:sec1_full}
    \end{subfigure}
\caption{Texture coordinate perturbations (texture sliding) reduce shape inference errors: ground truth (blue), prediction (orange).}
\label{fig:tshirt_example}
\end{figure}

We focus on the specific task of adding high-frequency wrinkles to virtual clothing, noting that the idea of learning a low-frequency embedding may be generalized to other tasks.
Because cloth wrinkles/folds are high-frequency features, existing deep neural networks (DNNs) trained to infer cloth shape tend to predict overly smooth meshes \cite{alldieck2019learning,danvevrek2017deepgarment,guan2012drape,gundogdu2019garnet,jin2018pixel,lahner2018deepwrinkles,natsume2019siclope,santesteban2019learning,wang2018learning}.
Rather than attempting to amend such errors directly, we perturb texture so that the rendered cloth mesh \textit{appears} to more closely match the ground truth.
See Figure \ref{fig:tshirt_example}.
Then given texture perturbations from at least two unique camera views, 3D geometry can be accurately reconstructed \cite{hartley1997triangulation} to recover high-frequency wrinkles.
Similarly, for AR/VR applications, correcting visual appearance from two views (one for each eye) is enough to allow the viewer to accurately discern 3D geometry.

Our proposed texture coordinate perturbations are highly dependent on the camera view.
Thus, we demonstrate that one can train a separate texture sliding neural network (TSNN) for each of a finite number of cameras laid out into an array and use nearby networks to interpolate results valid for any view enveloped by the array.
Although an approach similar in spirit might be pursued for various lighting conditions, this limitation is left as future work since there are a great deal of applications where the light is ambient/diffuse/non-directional/etc.
In such situations, this further complication may be ignored without significant repercussion. 

\section{Related Work}
\textbf{Cloth:}
While physically-based cloth simulation has matured as a field over the last few decades \cite{baraff1998large,baraff2003untangling,bridson2002robust,bridson2003simulation,selle2008robust}, data-driven methods are attractive for many applications.
There is a rich body of work in reconstructing cloth from multiple views or 3D scans, see \eg \cite{bradley2008markerless,franco2006visual,vlasic2008articulated}.
More recently, optimization-based methods have been used to generate higher resolution reconstructions \cite{huang2015hybrid,pons2017clothcap,wu2012full,yang2016estimation}.
Some of the most interesting work focuses on reconstructing the body and cloth separately \cite{bualan2008naked,neophytou2014layered,yang2018analyzing,zhang2017detailed}.

\textbf{Cloth and Learning:}
With advances in deep learning, one can aim to reconstruct 3D cloth meshes from single views.
A number of approaches reconstruct a joint cloth/body mesh from a single RGB image \cite{alldieck2019learning,alldieck2019tex2shape,natsume2019siclope,saito2019pifu}, RGB-D image \cite{yu2019simulcap}, or video \cite{alldieck2018detailed,alldieck2018video,habermann2019livecap,xu2018monoperfcap}.
To reduce the dimensionality of the output space, DNNs are often trained to predict the pose/shape parameters of human body models such as SCAPE \cite{anguelov2005scape} or SMPL \cite{loper2015smpl} (see also \cite{pavlakos2019expressive}).
\cite{alldieck2019learning,alldieck2018detailed,alldieck2018video} predict SMPL model parameters along with per-vertex offsets to add details, and \cite{alldieck2019tex2shape} refines the mesh using the network proposed in \cite{isola2017image}.
\cite{habermann2019livecap,natsume2019siclope,varol2018bodynet} leverage predicted pose information to infer shape.
Estimating shape from silhouettes given an RGB image has also been explored \cite{dibra2016hs,dibra2017human,natsume2019siclope}.
When only the garment shape is predicted, a number of recent works output predictions in UV space to represent geometric information as pixels \cite{danvevrek2017deepgarment,jin2018pixel,lahner2018deepwrinkles}, although others \cite{gundogdu2019garnet,santesteban2019learning} define loss functions directly in terms of the 3D cloth vertices.

\textbf{Wrinkles and Folds:}
Cloth realism can be improved by introducing wrinkles and folds.
In the graphics community, researchers have explored both procedural and data-driven methods for generating wrinkles \cite{de2010stable,guan2012drape,hahn2014subspace,muller2010wrinkle,rohmer2010animation,wang2010example}.
Other works add real-world wrinkles as a postprocessing step to improve smooth captured cloth: \cite{popa2009wrinkling} extracts the edges of cloth folds and then applies space-time deformations, \cite{robertini2014efficient} solves for shape deformations directly by optimizing over all frames of a video sequence.
Recently, \cite{lahner2018deepwrinkles} used a conditional Generative Adversarial Network \cite{mirza2014conditional} to generate normal maps as proxies for wrinkles on captured cloth.

\textbf{Geometry:}
More broadly, deep learning on 3D meshes falls under the umbrella of \textit{geometric deep learning}, which was coined by \cite{bronstein2017geometric} to characterize learning in non-Euclidean domains.
\cite{scarselli2008graph} was one of the earliest works in this area and introduced the notion of a Graph Neural Network (GNN) in relation to CNNs.
Subsequent works similarly extend the CNN architecture to graphs and manifolds \cite{boscaini2016learning,maron2017convolutional,masci2015geodesic,monti2017geometric}.
\cite{kostrikov2018surface} introduces a latent representation that explicitly incorporates the Dirac operator to detect principal curvature directions.
\cite{tan2018variational} trains a mesh generative model to generate novel meshes outside an original dataset.
Returning to the specific application of virtual cloth, \cite{jin2018pixel} embeds a non-Euclidean cloth mesh into a Euclidean pixel space, making it possible to directly use CNNs to make non-Euclidean predictions.

\textbf{Texture:}
In the computer graphics community, textures have historically been used to capture both geometric and material details lost by using simplified models \cite{foley1996computer,marschner2015fundamentals}, which is similar in spirit to our approach.
Though, to the best of our knowledge, we are the first to propose learning texture coordinate perturbations to facilitate the accurate reconstruction of lost geometric details.
For completeness, we briefly note a few works that use learning for texture synthesis and/or style transfer \cite{dumoulin2016learned,gatys2015neural,gatys2016image,gupta2017characterizing,johnson2016perceptual,sanakoyeu2018style}.

\section{Methods}
We define \textit{texture sliding} as the changing of texture coordinates on a per-camera basis such that any point which is visible from some stereo pair of cameras can be triangulated back to its ground truth position.
Other stereo reconstruction techniques can also be used in place of triangulation because the \textit{images} we generate are consistent with the ground truth geometry.
See \eg \cite{bradley2008accurate,hartley1997triangulation,seitz2006comparison}.

\subsection{Per-Vertex Discretization}
\begin{figure}[ht]
\centering
\begin{subfigure}[b]{0.49\linewidth}
    \includegraphics[width=\linewidth]{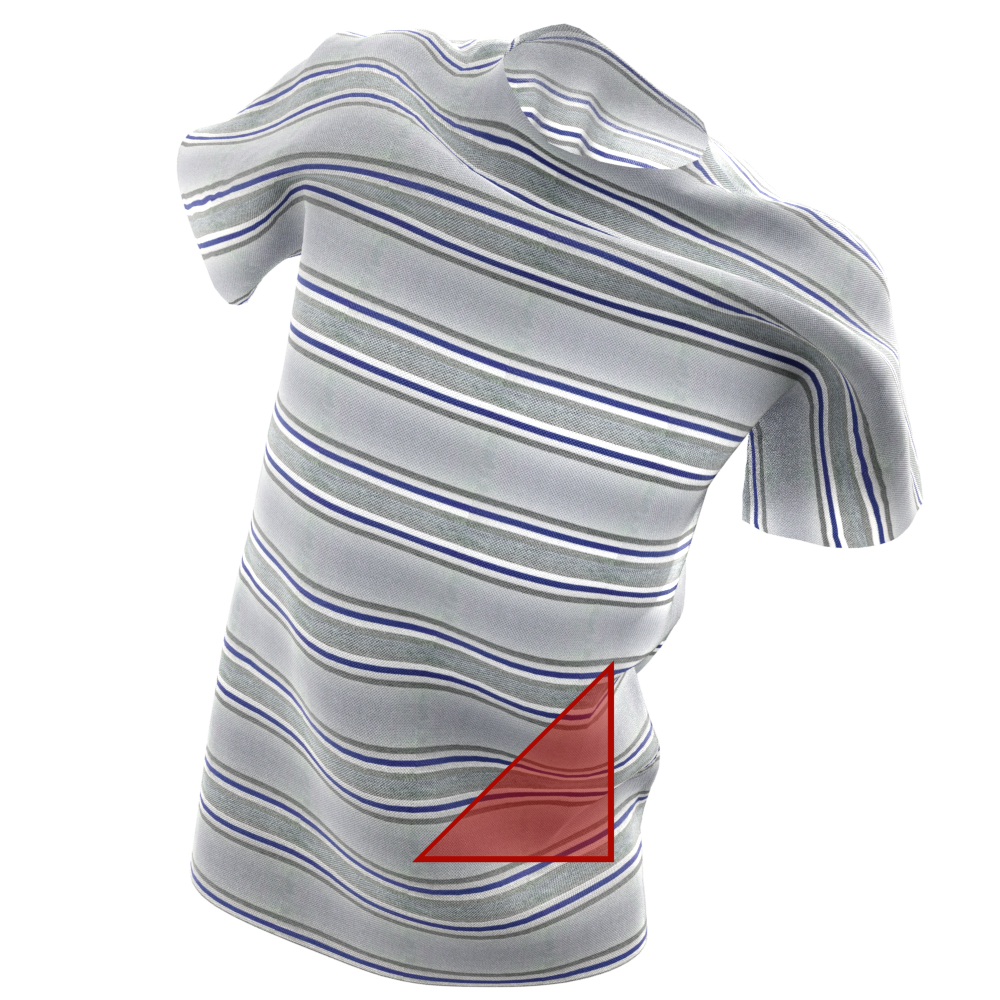}
    \label{fig:bigtriangle_0}
\end{subfigure}
\begin{subfigure}[b]{0.49\linewidth}
    \includegraphics[width=\linewidth]{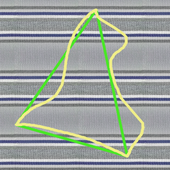}
    \label{fig:bigtriangle_1}
\end{subfigure}
\hfill
\caption{
Consider an extreme case, where the inferred cloth has a quite large triangle (shown in red).
That triangle should encompass the nonlinear texture region outlined in yellow (shown in pattern space).
Note: the yellow curve was generated by sampling the ground truth cloth's texture coordinates along the projected edges of the red triangle.
The linearity assumption implied by barycentric interpolation instead uses the region outlined in green.
}
\label{fig:bigtriangle}
\end{figure}

Since the cloth mesh is discretized into vertices and triangles, we take a per-vertex, not a per-point, approach to texture sliding.
Our proposed method (see Section \ref{ray}) computes per-vertex texture coordinates on the inferred cloth that match those of the ground truth as seen by the camera under consideration.
Then during 3D reconstruction, barycentric interpolation is used to find the subtriangle locations of the texture coordinates corresponding to ground truth cloth vertices.
This assumes linearity, which is only valid when the triangles are small enough to capture the inherent nonlinearities in a piecewise linear sense; moreover, folds and wrinkles can create significant nonlinearity.
See Figure \ref{fig:bigtriangle}.

\subsection{Occlusion Boundaries}
Accurate 3D reconstruction requires that a vertex of the ground truth mesh be visible from at least two cameras \textit{and} that camera projections of the vertex to the inferred cloth exist and are valid.
However, occlusions can derail these assumptions.

\begin{figure}[ht]
\centering
\begin{subfigure}[b]{0.49\linewidth}
    \includegraphics[width=\linewidth]{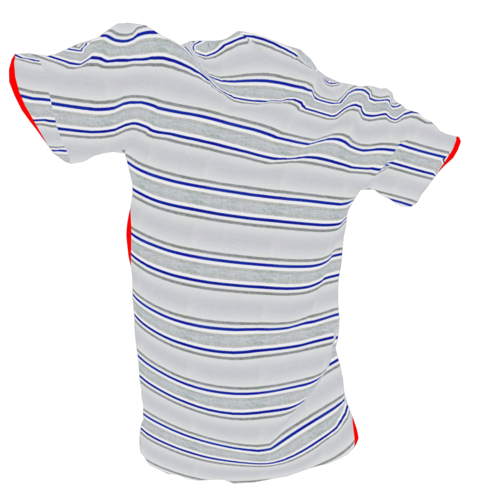}
    \caption{}
    \label{fig:occlusion_boundary_pd}
\end{subfigure}
\begin{subfigure}[b]{0.49\linewidth}
    \includegraphics[width=\linewidth]{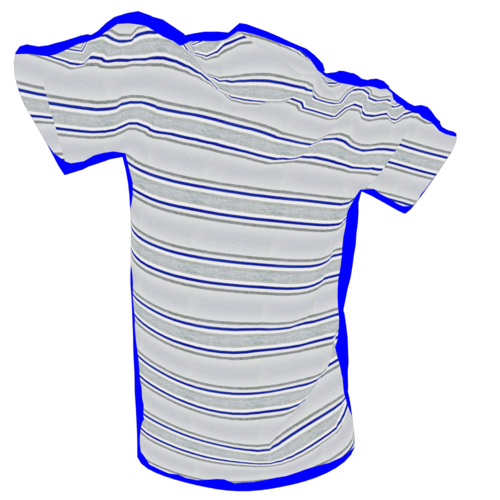}
    \caption{}
    \label{fig:occlusion_boundary_gt}
\end{subfigure}
\hfill
\caption{
The method discussed in Section \ref{ray} can fail near silhouettes of the inferred and ground truth cloth meshes, in which case smoothness assumptions are used (see Section \ref{diffusion}).
In (a), inferred triangles with at least one vertex falling outside the silhouette of the ground truth mesh are colored red. In (b), ground truth triangles with at least one vertex falling outside the silhouette of the inferred mesh are colored blue.
}
\label{fig:occlusion_boundary}
\end{figure}

First, consider things from the standpoint of the inferred cloth.
For a given camera view, some inferred cloth triangles will not contain any visible pixels, and we denote a vertex as occluded when none of its incident triangles contain any visible pixels.
Although we do not assign perturbed texture coordinates to occluded vertices (\ie they keep their original texture coordinates, or a perturbation of zero), we do aim to keep the texture coordinate perturbation function smooth (see Section \ref{diffusion}).
In addition, there will be so called non-occluded vertices in the inferred cloth that do not project to visible pixels of the ground truth cloth.
This often occurs near silhouette boundaries where the inferred cloth silhouette is sometimes wider than the ground truth cloth silhouette.
These vertices are also treated as occluded, similar to those around the back side of the cloth behind the silhouette, essentially treating some extra vertices near occlusion boundaries as also being occluded.
See Figure \ref{fig:occlusion_boundary_pd}.

Next, consider things from the standpoint of the ground truth cloth.
For example, consider the case where all the cameras are in the front, and vertices on the back side of the ground truth cloth are not visible from any camera.
The best one can do in reconstructing these occluded vertices is to use the inferred cloth vertex positions; however, care should be taken near occlusion boundaries to smoothly taper between our texture sliding 3D reconstruction and the inferred cloth prediction.
A simple approach is to extrapolate/smooth the geometric difference between our texture sliding 3D reconstruction and the inferred cloth prediction to occluded regions of the mesh.
Once again, the definition of occluded vertices needs to be broadened for silhouette consideration.
Not only will vertices not visible from at least two cameras have to be considered occluded, but vertices that don't project to the interior of an inferred cloth triangle with \textit{valid} texture coordinate perturbations will also have to be considered occluded.
See Figure \ref{fig:occlusion_boundary_gt}.

\section{Dataset Generation} \label{data_generation}
Let $C = \{X,T\}$ be a cloth triangulated surface with $n$ vertices $X \in \mathbb{R}^{3n}$ and texture coordinates $T \in \mathbb{R}^{2n}$.
We assume that mesh connectivity remains fixed throughout.
The ground truth cloth mesh $C_G(\theta) = \{X_G(\theta),T_G\}$ depends on the pose $\theta$.
Given a pre-trained DNN (we use the network from \cite{jin2018pixel}), the inferred cloth $C_N(\theta)=\{X_N(\theta),T_G\}$ is also a function of the pose $\theta$.
Our objective is to replace the ground truth texture coordinates $T_G$ with perturbed texture coordinates $T_N(\theta,v)$, \ie to compute $C_N'(\theta,v) = \{X_N(\theta), T_N(\theta,v)\}$ where $T_N(\theta,v)$ depends on both the pose $\theta$ and the view $v$.
Even though $T_N(\theta,v)$ is in principle valid for all $v$ using interpolation (see Section \ref{multiview_results}), training data $T_N(\theta,v_p)$ is only required for a finite number of camera views $v_p$.
For each camera $p$, we also only require training data for finite number of poses $\theta_k$, \ie we require $T_N(\theta_k,v_p)$, which is computed from $T_G$ using $X_G(\theta_k)$, $X_N(\theta_k)$, and $v_p$.

\subsection{Texture Coordinate Projection} \label{ray}
\begin{figure}[ht]
\centering
\includegraphics[width=\linewidth]{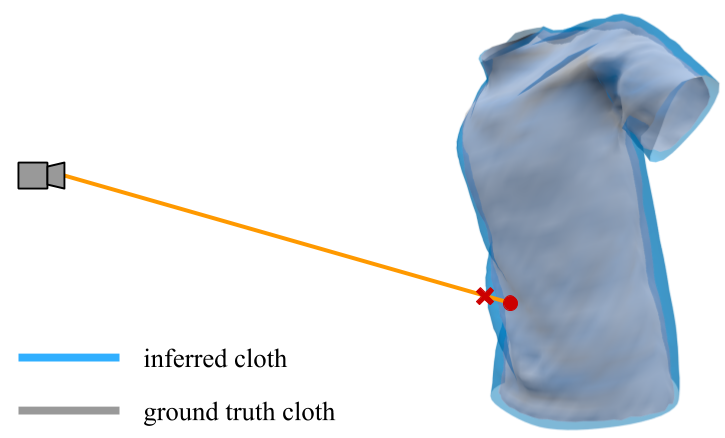}
\linebreak
\caption{Illustration of the ray intersection method for transferring texture coordinates to the inferred cloth from the ground truth cloth. Texture coordinates for the inferred cloth vertex (red cross) are interpolated from the ground truth mesh to the point of ray intersection (red circle).}
\label{fig:ray_intersection}
\end{figure}

We project texture coordinates to the inferred cloth vertices $X_N(\theta_k)$ from the ground truth cloth mesh $C_G(\theta_k)$ using ray intersection.
For each inferred cloth vertex in $X_N(\theta_k)$, we cast a ray from camera $p$'s aperture through the vertex
and find the first intersection with the ground truth mesh $C_G(\theta_k)$; subsequently, $T_G$ is barycentrically interpolated to the point of intersection and assigned to the inferred cloth vertex as its $T_N(\theta_k,v_p)$ value.
See Figure \ref{fig:ray_intersection}.
Rays are only cast for inferred cloth vertices that have at least one incident triangle with a nonzero area subregion visible to camera $p$.
Also, a ground truth texture coordinate value is only assigned to an inferred cloth vertex when the point of intersection with the ground truth mesh is visible to camera $p$.
We store and learn texture coordinate displacements $d_{v_p}(\theta_k) = T_N(\theta_k,v_p) - T_G$.
After this procedure, any remaining vertices of the inferred cloth that have not been assigned $d_{v_p}(\theta_k)$ values are treated as occluded and handled via smoothness considerations as discussed in Section \ref{diffusion}.

\subsection{Occlusion Handling} \label{diffusion}
Some vertices of the inferred cloth mesh remain unassigned with $d_{v_p}(\theta_k)=0$ after executing the algorithm outlined in Section \ref{ray}.
This creates a discontinuity in $d_{v_p}(\theta_k)$ which excites high frequencies that require a more complex network architecture to capture.
In order to alleviate demands on the network, we smooth $d_{v_p}(\theta_k)$ as follows.
First, we use the Fast Marching Method on triangulated surfaces \cite{kimmel1998computing} to generate a signed distance field.
Then, we extrapolate $d_{v_p}(\theta_k)$ normal to the distance field into the unassigned region, see \eg \cite{osher2002level}.
Finally, a bit of averaging is used to provide smoothness, while keeping the assigned values of $d_{v_p}(\theta_k)$ unchanged.
Alternatively, one could solve a Poisson equation as in \cite{cong2015fully} while using the assigned $d_{v_p}(\theta_k)$ as Dirichlet boundary conditions.

\section{Network Architecture} \label{training}
A separate texture sliding neural network (TSNN) is trained for each camera $p$; thus, we drop the $v_p$ notation in this section.
The loss is defined over all poses $\theta_k$ in the training set
\begin{equation} \label{eq:loss}
\mathcal{L} = \sum_{\theta_k} \left\lVert d(\theta_k) - \hat d(\theta_k) \right\rVert_2
\end{equation}
to minimize the difference between the desired displacements $d(\theta_k)$ and predicted displacements $\hat d(\theta_k)$.
The inferred cloth data we chose to correct are predictions of the T-shirt meshes from \cite{jin2018pixel}, each of which contains about 3,000 vertices.
The dataset spans about 10,000 different poses generated from a scanned garment using physically-based simulation.
To improve the resolution, we up-sampled each cloth mesh by subdividing each triangle into four subtriangles.
Notably, our texture sliding approach can be used to augment the results of any dataset for which ground truth and inferred training examples are available.
Moreover, it is trivial to increase the resolution of any such dataset simply by subdividing triangles.
Note that perturbations of the subdivided geometry are unnecessary, as we merely desire more sample points (to address Figure \ref{fig:bigtriangle}).
Finally, we applied an 80-10-10 training-validation-test set split.

\begin{figure}[ht]
\centering
\includegraphics[width=\linewidth]{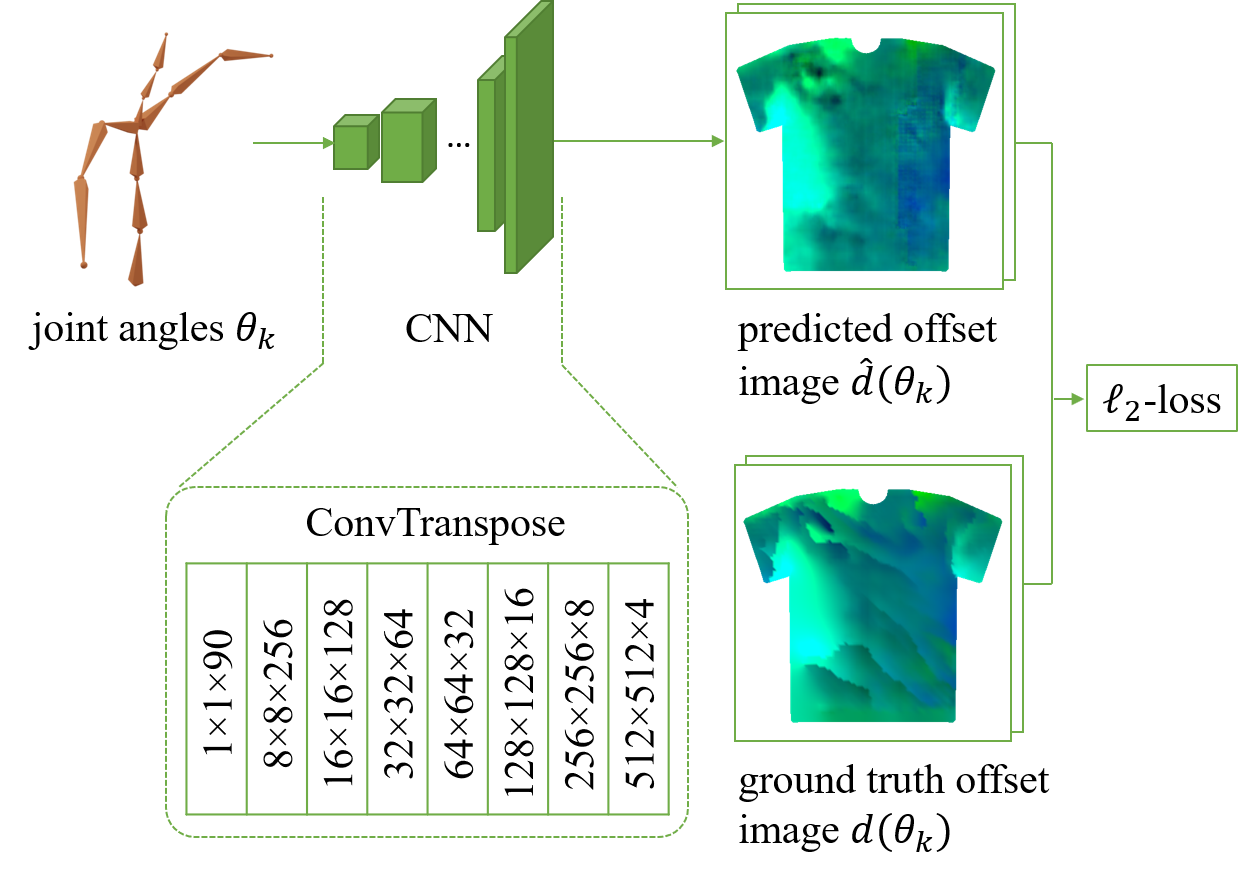}
\caption{Texture sliding neural network (TSNN) architecture.}
\label{fig:network}
\end{figure}

Similar to \cite{jin2018pixel}, the displacements $d(\theta_k)$ are stored as pixel-based cloth images for the front and back sides of the T-shirt, though we still output per-vertex texture coordinate displacements in UV space.
See Figure \ref{fig:network} for an overview of the network architecture.
Given input joint transformation matrices of shape $1\times1\times90$, TSNN applies a series of transpose convolution, batch normalization, and ReLU activation layers to upsample the input to $512\times512\times4$.
The first two dimensions of the output tensor represent the predicted displacements for the front side of the T-shirt, and the remaining two dimensions represent those for the back side.

\section{Experiments}
In Section \ref{gt_results}, we quantify the data generation approach of Section \ref{data_generation} and highlight the advantages of mesh subdivision for up-sampling.
In Section \ref{net_results}, we evaluate the predictions made by our trained texture sliding neural network (TSNN).
In Section \ref{multiview_results}, we demonstrate the interpolation of texture sliding results to novel views between a finite number of cameras.
Finally, in Section \ref{reconstruction_results}, we use multi-view texture sliding to reconstruct 3D geometry.
        
\subsection{Dataset Generation and Evaluation} \label{gt_results}
We aim to have the material coordinates of the cloth be in the correct locations as viewed by multiple cameras, so that the material can be accurately 3D reconstructed with point-wise accuracy.
As such, our error metric is a bit more stringent than that commonly used because our aim is to reproduce the actual material behavior, not merely to mimic its look (\eg, by perturbing normal vectors to create shading consistent with wrinkles in spite of the cloth being smooth, as in \cite{lahner2018deepwrinkles}).
In order to elucidate this, consider a two-step approach where one first approximates a smooth cloth mesh and then perturbs that mesh to add wrinkles (similar to \cite{santesteban2019learning}).
In order to preserve area and achieve the correct material behavior, material in the vicinity of a newly forming wrinkle should slide laterally towards that wrinkle as it is formed.
Merely non-physically stretching the material in order to create a wrinkle may look plausible, but does not admit the  correct material behavior.
In fact, the texture would be unrealistically stretched as well, although this is less apparent visually when using simple textures.

\begin{figure}[ht]
\centering
\begin{subfigure}[b]{0.32\linewidth}
    \includegraphics[width=\linewidth]{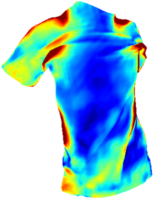}
    \caption{}
    \label{fig:gt_examples_a}
\end{subfigure}
\begin{subfigure}[b]{0.32\linewidth}
    \includegraphics[width=\linewidth]{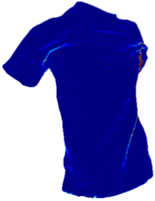}
    \caption{}
    \label{fig:gt_examples_b}
\end{subfigure}
\begin{subfigure}[b]{0.32\linewidth}
    \includegraphics[width=\linewidth]{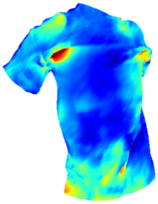}
    \caption{}
    \label{fig:gt_examples_c}
\end{subfigure}
\hfill
\caption{Per-pixel texture coordinate errors before (a) and after (b) applying texture sliding to the inferred cloth output by the network of \cite{jin2018pixel}. The result of a two-step process (c) may well match the ground truth in a visual sense, whilst still having quite large errors in material coordinates. Blue $=0$, red $\geq 0.04$.}
\label{fig:gt_examples}
\end{figure}

\begin{figure}[ht]
\centering
\begin{subfigure}[b]{0.32\linewidth}
    \includegraphics[width=\linewidth]{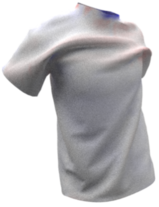}
    \caption{}
    \label{fig:gt_stress_a}
\end{subfigure}
\begin{subfigure}[b]{0.32\linewidth}
    \includegraphics[width=\linewidth]{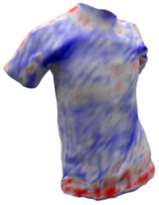}
    \caption{}
    \label{fig:gt_stress_b}
\end{subfigure}
\begin{subfigure}[b]{0.32\linewidth}
    \includegraphics[width=\linewidth]{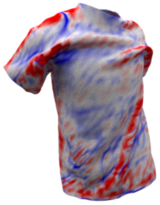}
    \caption{}
    \label{fig:gt_stress_c}
\end{subfigure}
\hfill
\caption{Local compression (blue) and extension (red) energies for a sample pose, comparing the ground truth cloth (a), the inferred cloth (b), and the result of a two-step process (c). In spite of the cloth mesh in (c) bearing visual resemblance to the ground truth in (a), it still has quite erroneous deformation energies.}
\label{fig:gt_stress}
\end{figure}

Since texture coordinates provide a proxy surface parameterization for material coordinates, we measure texture coordinate errors in a per-pixel fashion comparing between the ground truth and inferred cloth at the center of each pixel.
Figure \ref{fig:gt_examples_a} shows results typical for cloth inferred using the network from \cite{jin2018pixel}, and Figure \ref{fig:gt_examples_b} shows the highly improved results obtained on the same inferred geometry using our texture sliding approach (with 1 level of subdivision).
Note that the vast majority of the errors in Figure \ref{fig:gt_examples_b} occur near the wrinkles where the nonlinearities illustrated in Figure \ref{fig:bigtriangle} are most prevalent.
In Figure \ref{fig:gt_examples_c}, we deform the vertices of the inferred cloth mesh so that they lie exactly on the ground truth mesh in order to mimic a two-step approach (as discussed above).
Note how our error metric captures the still rather large errors in the material coordinates (and thus cloth vertex positions) in spite of the mesh in Figure \ref{fig:gt_examples_c} appearing to have the same wrinkles and folds as the ground truth mesh.
Figure \ref{fig:gt_stress} compares the local compression and extension energies of the ground truth mesh (Figure \ref{fig:gt_stress_a}), the inferred cloth mesh (Figure \ref{fig:gt_stress_b}), and the result of this two-step process (Figure \ref{fig:gt_stress_c}).
In spite of the untextured mesh in Figure \ref{fig:gt_stress_c} bearing visual similarity to the ground truth in Figure \ref{fig:gt_stress_a}, it still has rather large errors in deformation energy.

\begin{figure}[ht]
\centering
\begin{subfigure}[b]{0.19\linewidth}
    \includegraphics[width=\linewidth]{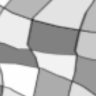}
    \caption{}
    \label{fig:subdivision_a}
\end{subfigure}
\begin{subfigure}[b]{0.19\linewidth}
    \includegraphics[width=\linewidth]{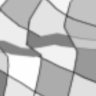}
    \caption{}
    \label{fig:subdivision_b}
\end{subfigure}
\begin{subfigure}[b]{0.19\linewidth}
    \includegraphics[width=\linewidth]{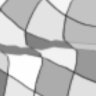}
    \caption{}
    \label{fig:subdivision_c}
\end{subfigure}
\begin{subfigure}[b]{0.19\linewidth}
    \includegraphics[width=\linewidth]{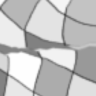}
    \caption{}
    \label{fig:subdivision_d}
\end{subfigure}
\begin{subfigure}[b]{0.19\linewidth}
    \includegraphics[width=\linewidth]{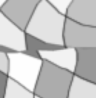}
    \caption{}
    \label{fig:subdivision_e}
\end{subfigure}
\hfill
\caption{
As the inferred cloth mesh (a) is subdivided, texture sliding (b-d) moves the appearance of the inferred mesh closer to the ground truth (e).
}
\label{fig:subdivision}
\end{figure}

Figure \ref{fig:subdivision} illustrates the efficacy of subdividing the cloth mesh to get more samples for texture sliding.
The particular ground truth cloth wrinkle shown in Figure \ref{fig:subdivision_e} is not captured by the inferred cloth geometry shown in Figure \ref{fig:subdivision_a}.
The texture sliding result shown in Figure \ref{fig:subdivision_b} better represents the ground truth cloth.
Figures \ref{fig:subdivision_c} and \ref{fig:subdivision_d} show how subdividing the inferred cloth mesh one and two times (respectively) progressively alleviates errors emanating from the linearity assumption illustrated in Figure \ref{fig:bigtriangle}.
Table \ref{tab:gt_mse} shows quantitative results comparing the inferred cloth to texture sliding with and without subdivision.

\begin{table}[ht]
\begin{center}
\begin{tabular}{|c|c|}
\hline
Method & SqrtMSE ($\times 10^{-3}$)\\
\hline\hline
Jin et al. \cite{jin2018pixel} & 24.496 $\pm$ 6.9536 \\
TS & 5.2662 $\pm$ 2.2320 \\
TS $+$ subdivision & 3.5645 $\pm$ 1.6617 \\
\hline
\end{tabular}
\end{center}
\caption{Per-pixel square root of mean squared error (SqrtMSE) for the entire dataset.}
\label{tab:gt_mse}
\end{table}

\subsection{Network Training and Inference} \label{net_results}
\begin{figure}[ht]
\centering
\begin{subfigure}[b]{0.32\linewidth}
    \includegraphics[width=\linewidth]{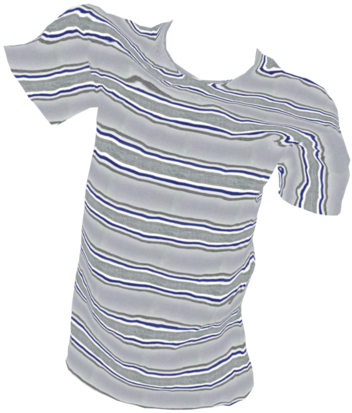}
    \caption{$\hat C_N'$}
\end{subfigure}
\begin{subfigure}[b]{0.32\linewidth}
    \includegraphics[width=\linewidth]{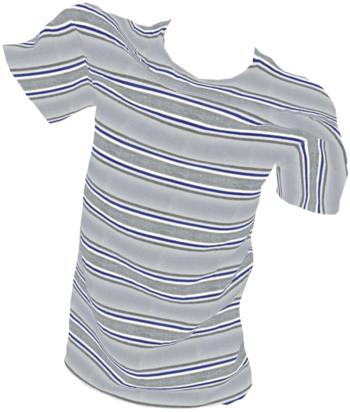}
    \caption{$C_N'$}
\end{subfigure}
\begin{subfigure}[b]{0.32\linewidth}
    \includegraphics[width=\linewidth]{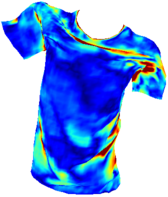}
    \caption{$Error(\hat C_N', C_G)$}
\end{subfigure}
\hfill
\caption{A typical test set example prediction. The per-pixel errors are shown in (c) (blue $=0$, red $\geq 0.04$).}
\label{fig:net_examples}
\end{figure}

\begin{figure}[ht]
\centering
\begin{subfigure}[b]{0.2\linewidth}
    \includegraphics[width=\linewidth]{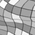}
    \caption{}
    \label{fig:subdivision_net_a}
\end{subfigure}
\begin{subfigure}[b]{0.2\linewidth}
    \includegraphics[width=\linewidth]{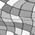}
    \caption{}
    \label{fig:subdivision_net_b}
\end{subfigure}
\begin{subfigure}[b]{0.2\linewidth}
    \includegraphics[width=\linewidth]{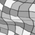}
    \caption{}
    \label{fig:subdivision_net_c}
\end{subfigure}
\begin{subfigure}[b]{0.2\linewidth}
    \includegraphics[width=\linewidth]{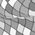}
    \caption{}
    \label{fig:subdivision_net_d}
\end{subfigure}
\hfill
\caption{
The results of the TSNN before (b) and after (c) subdivision, as compared to the ground truth (d).
In spite of Table 2, some wrinkles are better resolved by the TSNN after subdivision.
The inferred mesh with ground truth texture coordinates is shown in (a).
}
\label{fig:subdivision_net}
\end{figure}

The network was trained using the Adam optimizer \cite{kingma2014adam} with a $10^{-3}$ learning rate in PyTorch \cite{paszke2017automatic}.
As mentioned earlier, we subdivided the mesh triangles once.
Figure \ref{fig:net_examples} shows a typical prediction on a test set example, including the per-pixel errors in predicted texture coordinates.
While the TSNN is able to recover the majority of the shirt, it struggles near wrinkles.
Figure \ref{fig:subdivision_net} highlights a particular wrinkle comparing the inferred cloth (Figure \ref{fig:subdivision_net_a}) and the results of the TSNN before (Figure \ref{fig:subdivision_net_b}) and after (Figure \ref{fig:subdivision_net_c}) subdivision to the ground truth (Figure \ref{fig:subdivision_net_d}).
Table \ref{tab:net_mse} shows quantitative results comparing the inferred cloth to TSNN results with and without subdivision.

\begin{table}[ht]
\begin{center}
\begin{tabular}{|c|c|}
\hline
Network & SqrtMSE ($\times 10^{-3}$)\\
\hline\hline
Jin et al. \cite{jin2018pixel} & 24.871 $\pm$ 7.0613 \\
TSNN & 13.335 $\pm$ 4.2924 \\
TSNN + subdivision & 13.591 $\pm$ 4.5194 \\
\hline
\end{tabular}
\end{center}
\caption{Per-pixel SqrtMSE for the test set. Inspite of Table \ref{tab:gt_mse} demonstrating that subdivision improves the ground truth TS data, the improvements are not uniformly realized by the TSNN (which we discuss in the appendix).}
\label{tab:net_mse}
\end{table}

\begin{figure*}[t]
\centering
\begin{subfigure}[b]{0.16\linewidth}
    \includegraphics[width=\linewidth]{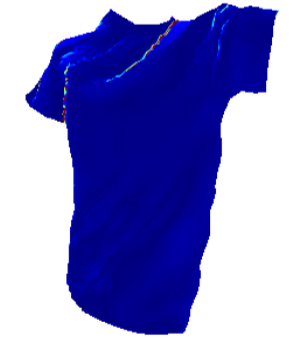}
\end{subfigure}
\begin{subfigure}[b]{0.16\linewidth}
    \includegraphics[width=\linewidth]{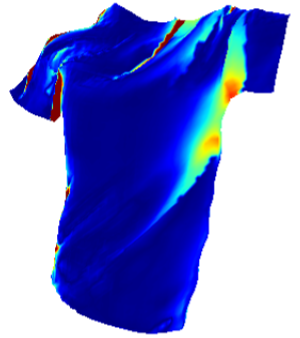}
\end{subfigure}
\begin{subfigure}[b]{0.16\linewidth}
    \includegraphics[width=\linewidth]{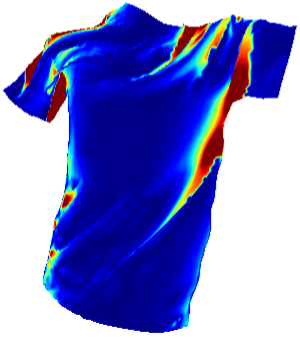}
\end{subfigure}
\begin{subfigure}[b]{0.16\linewidth}
    \includegraphics[width=\linewidth]{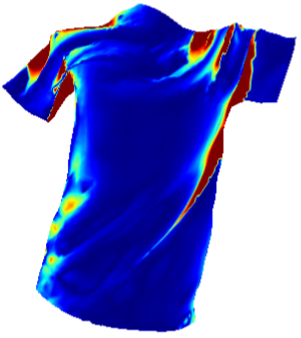}
\end{subfigure}
\begin{subfigure}[b]{0.16\linewidth}
    \includegraphics[width=\linewidth]{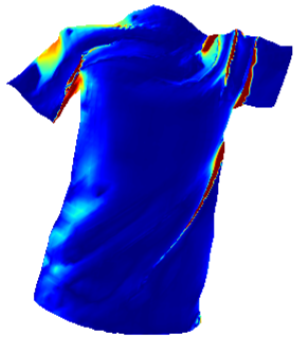}
\end{subfigure}
\begin{subfigure}[b]{0.16\linewidth}
    \includegraphics[width=\linewidth]{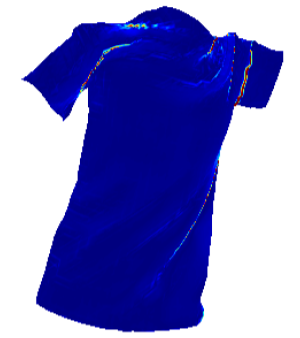}
\end{subfigure}
\hfill
\begin{subfigure}[b]{0.16\linewidth}
    \includegraphics[width=\linewidth]{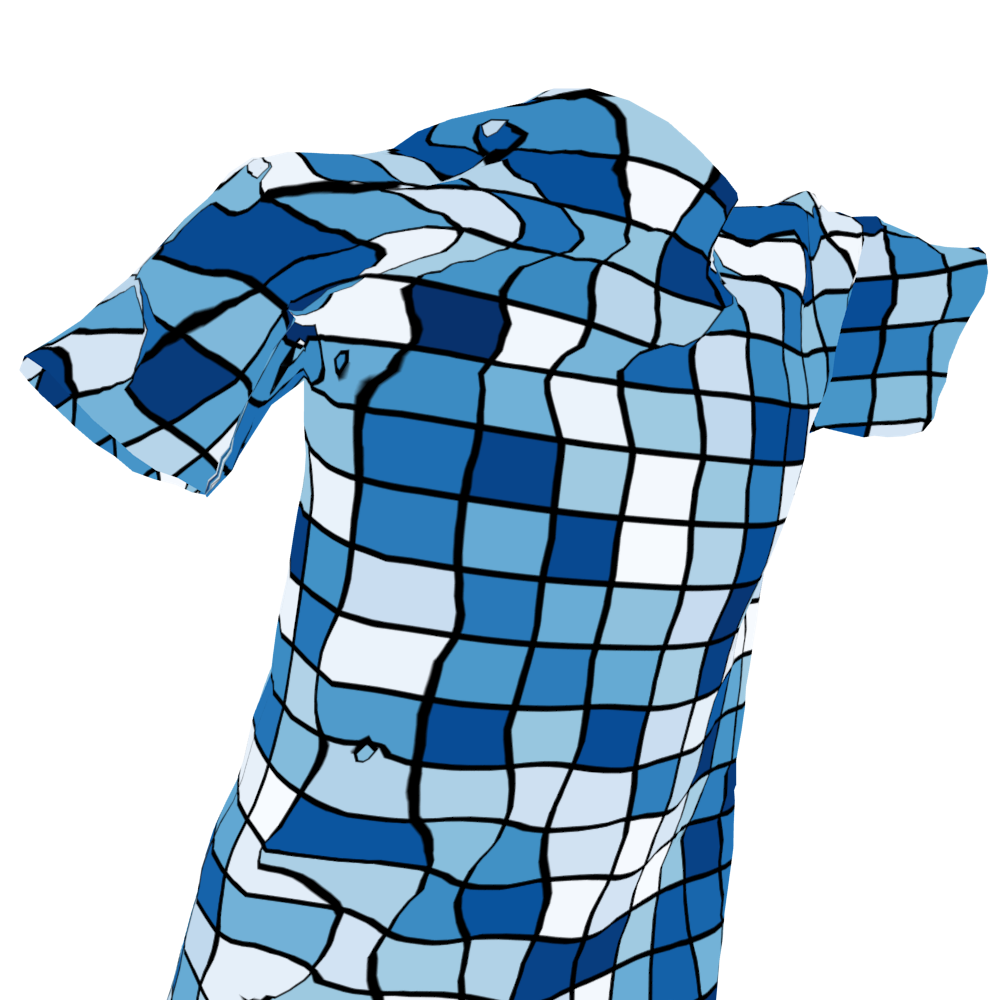}
\end{subfigure}
\begin{subfigure}[b]{0.16\linewidth}
    \includegraphics[width=\linewidth]{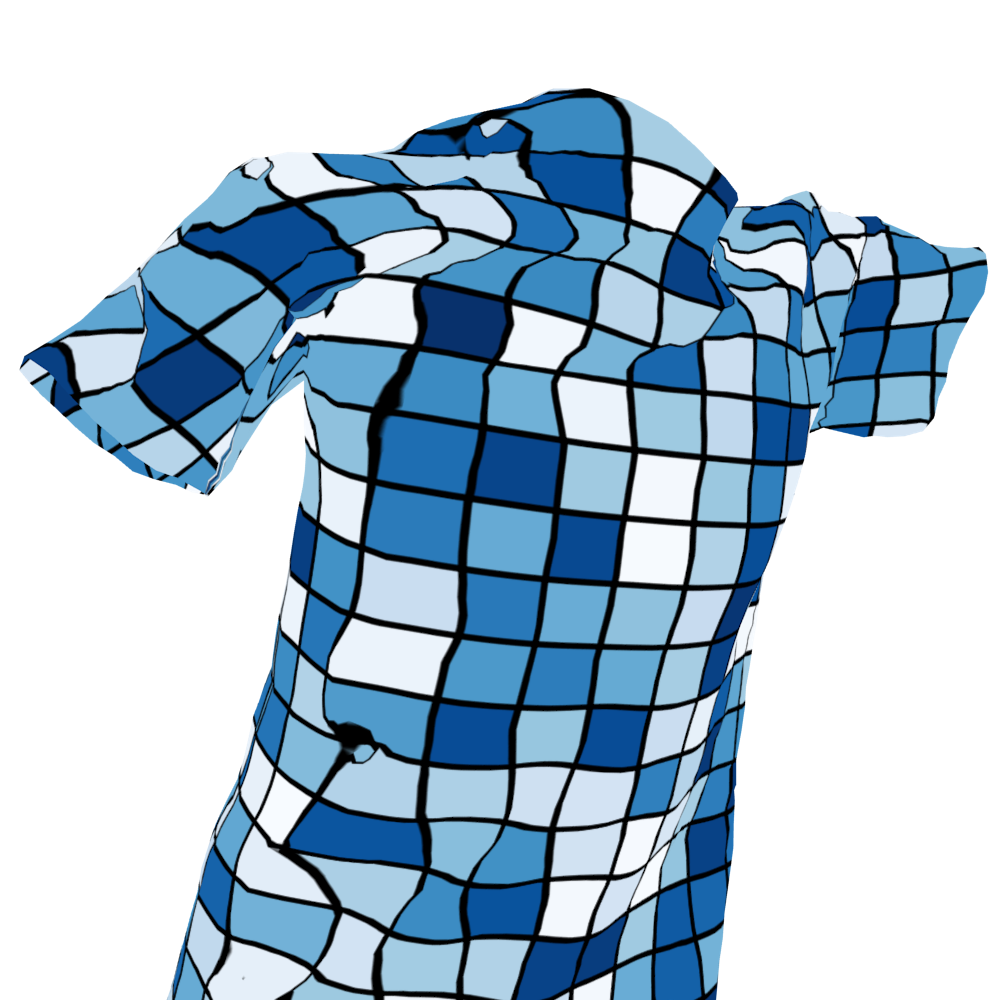}
\end{subfigure}
\begin{subfigure}[b]{0.16\linewidth}
    \includegraphics[width=\linewidth]{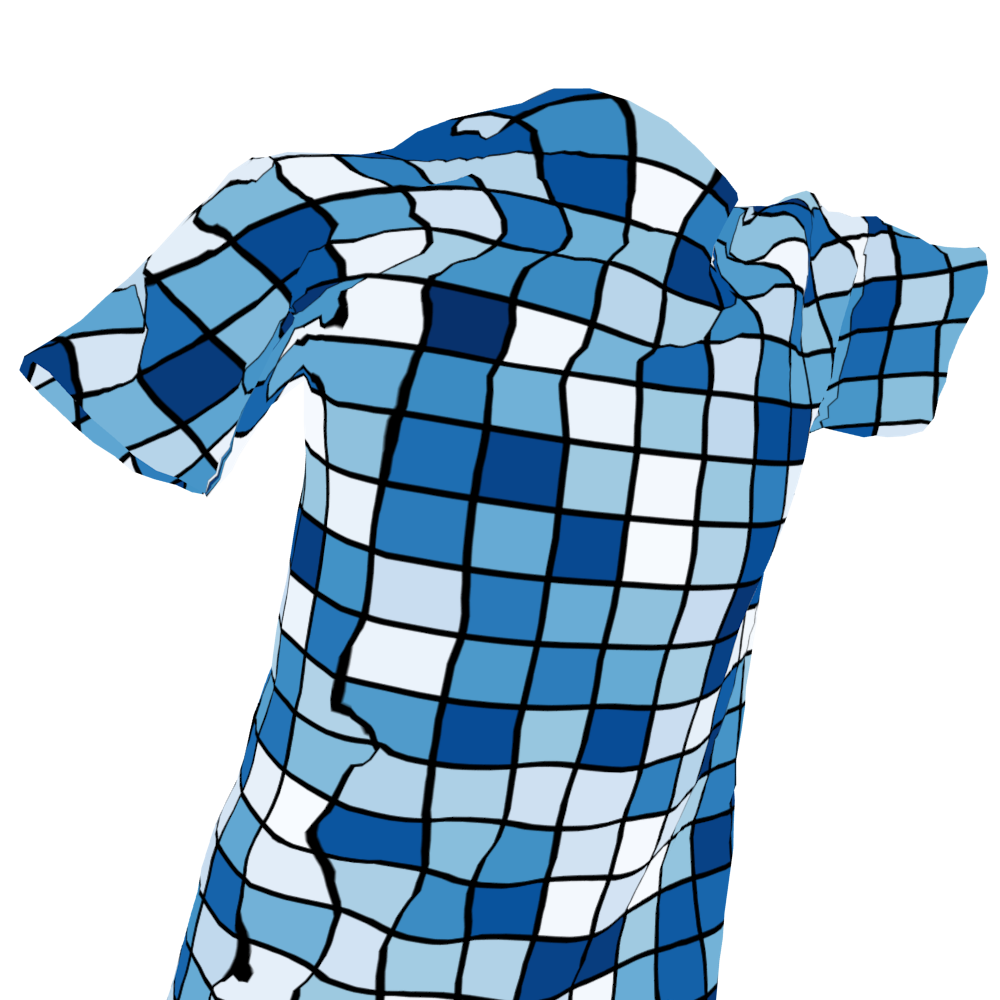}
\end{subfigure}
\begin{subfigure}[b]{0.16\linewidth}
    \includegraphics[width=\linewidth]{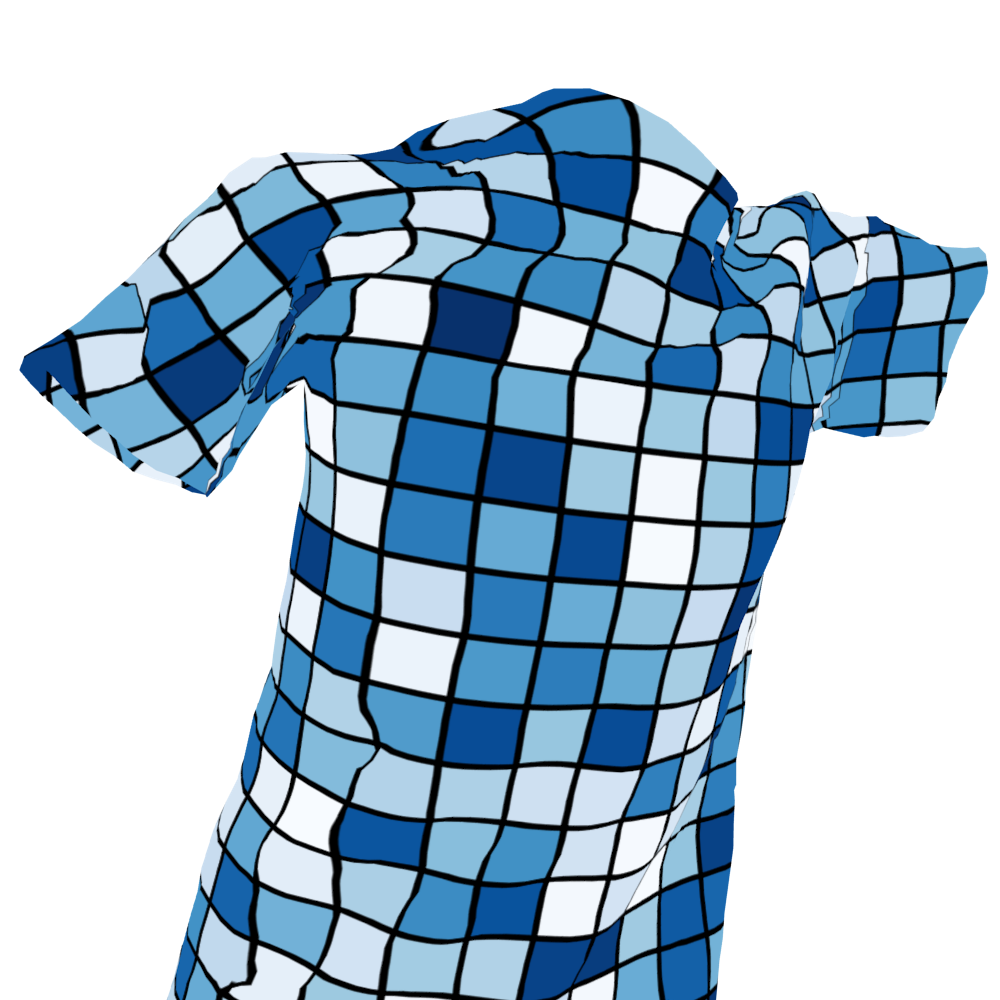}
\end{subfigure}
\begin{subfigure}[b]{0.16\linewidth}
    \includegraphics[width=\linewidth]{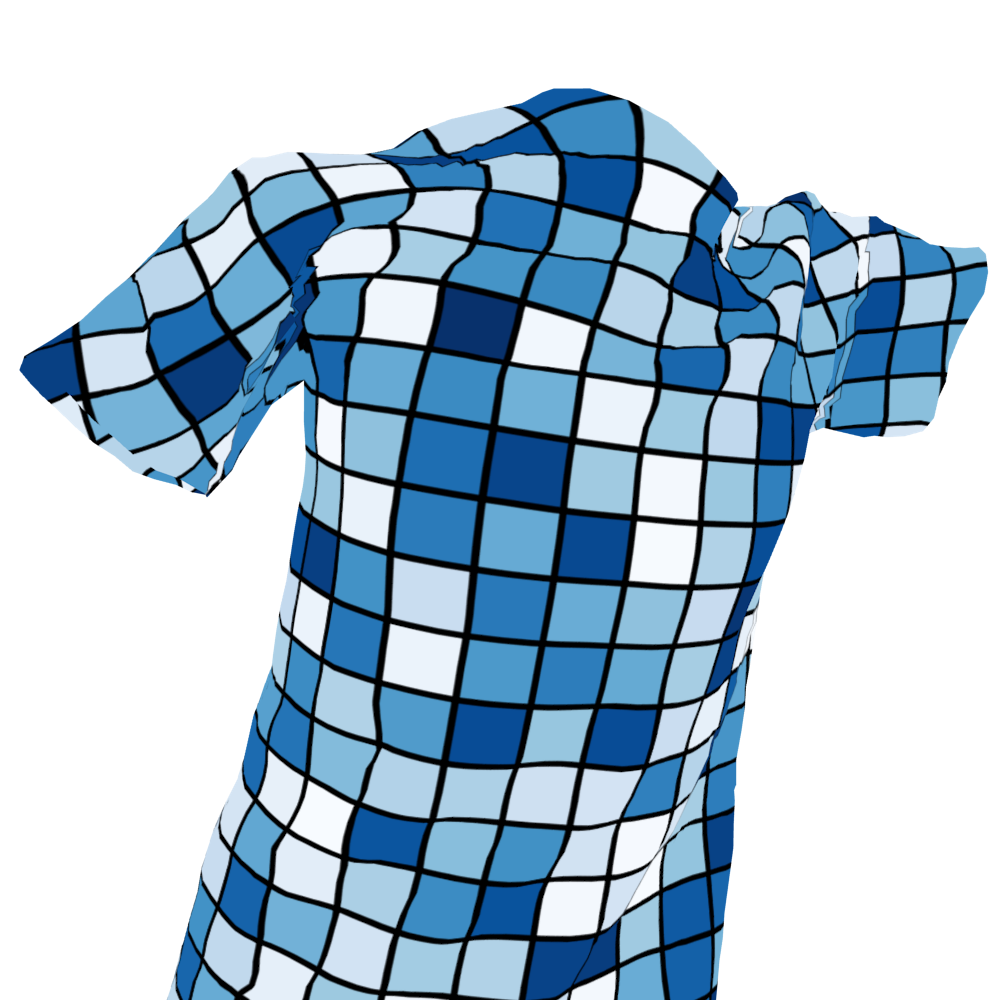}
\end{subfigure}
\begin{subfigure}[b]{0.16\linewidth}
    \includegraphics[width=\linewidth]{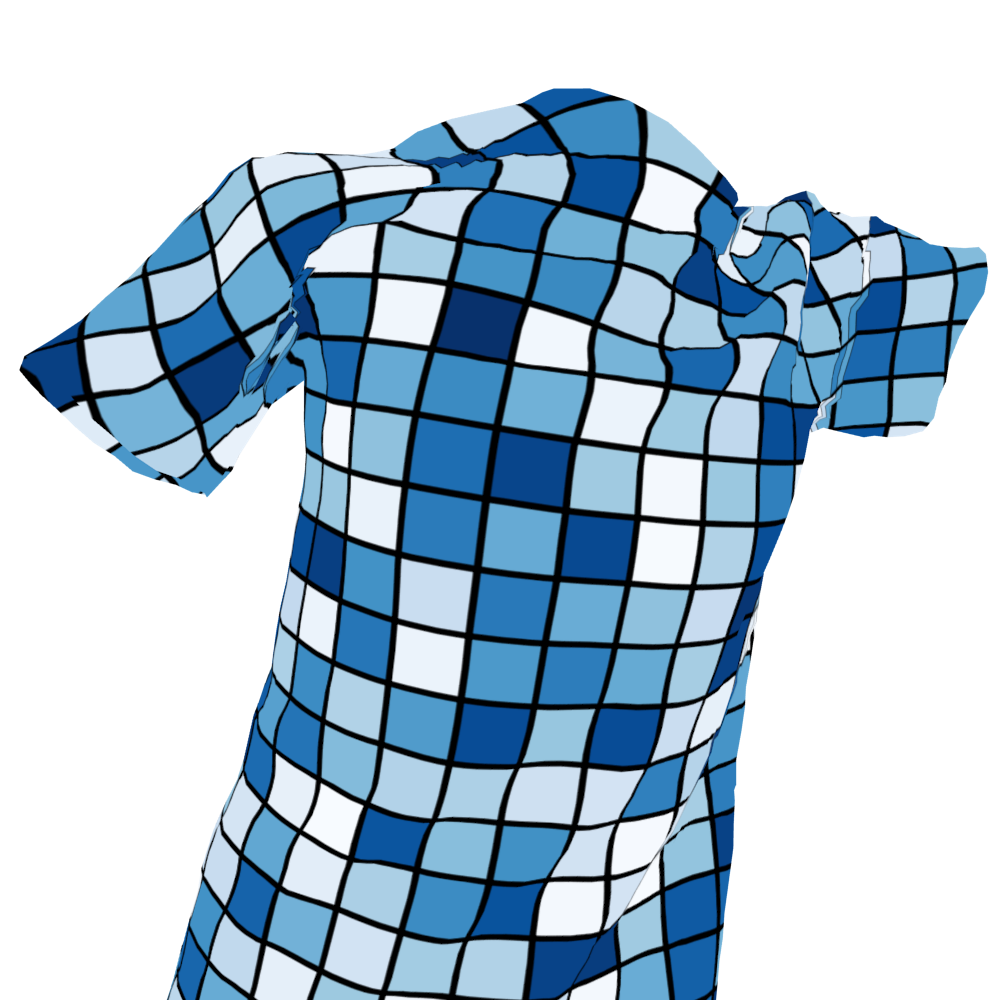}
\end{subfigure}
\hfill
\caption{Given two camera views (far left and far right images), texture sliding can be linearly interpolated to novel views between them. The top row shows per-pixel errors (blue $=0$, red $\geq 0.04$), and the bottom row shows the cloth from a \textit{fixed} front-facing view to illustrate how the interpolated texture changes as a function of the chosen novel view.}
\label{fig:multi_examples}
\end{figure*}

\subsection{Interpolating to Novel Views} \label{multiview_results}
\begin{figure}[ht]
\centering
\includegraphics[width=0.93\linewidth]{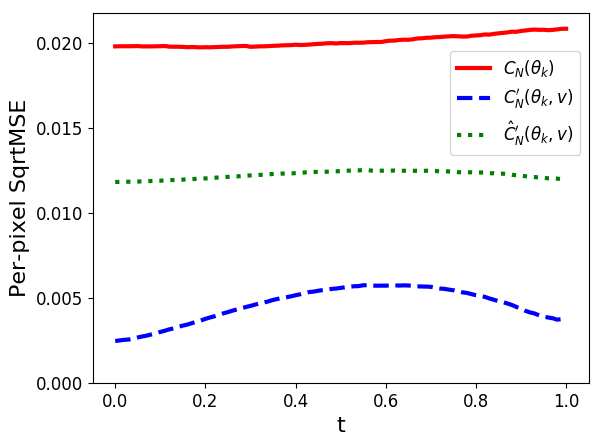}
\caption{Per-pixel SqrtMSE for interpolating between two cameras (using a test set example). Note that the inferred cloth does not use any view based information, but that our error metric does depend on the view.}
\label{fig:multi_plot}
\end{figure}

\begin{figure}[ht]
\centering
\begin{subfigure}[b]{\linewidth}
    \includegraphics[width=\linewidth]{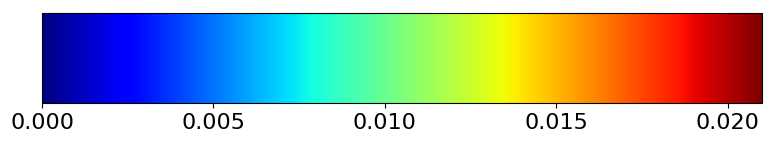}
\end{subfigure}
\begin{subfigure}[b]{0.3\linewidth}
    \includegraphics[width=\linewidth]{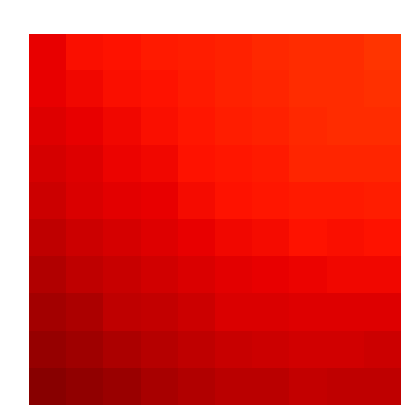}
    \caption{$C_N(\theta_k)$}
\end{subfigure}
\begin{subfigure}[b]{0.3\linewidth}
    \includegraphics[width=\linewidth]{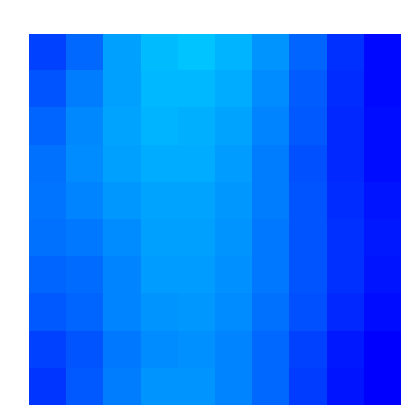}
    \caption{$C_N'(\theta_k,v)$}
\end{subfigure}
\begin{subfigure}[b]{0.3\linewidth}
    \includegraphics[width=\linewidth]{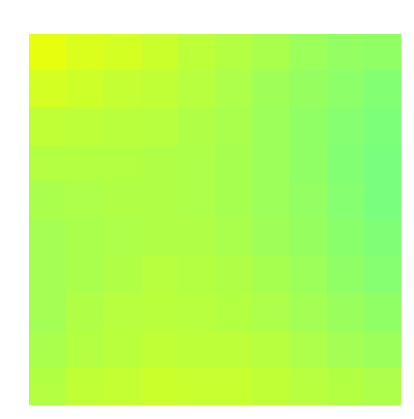}
    \caption{$\hat C_N'(\theta_k,v)$}
\end{subfigure}
\hfill
\caption{Per-pixel SqrtMSE for interpolating between four cameras (one at each corner of the square). The pose $\theta_k$ is the same as in Figure \ref{fig:multi_plot}, which plots the values along the bottom edge of the square.} 
\label{fig:multi_2d}
\end{figure}

Given a finite number of camera views $v_p$, one can specify a new view enveloped by the array using a variety of interpolation methods.
For the sake of demonstration, we take a simple approach assuming that one can interpolate via $v = \sum_p w_p v_p$, and then use these same weights to compute
\begin{equation} \label{eq:multi_interpolation}
T_N(\theta_k,v) = \sum_p w_p T_N(\theta_k,v_p)
\end{equation}
This same equation is also used for $\hat T_N(\theta_k, v)$. 
Figure \ref{fig:multi_examples} shows the results obtained by linearly interpolating between two camera views.
Note how the largest errors appear near areas occluded by wrinkles, where one (or both) of the cameras has no valid texture sliding results and instead uses the inferred cloth textures.
This can be alleviated by using more cameras placed closer together.
Figure \ref{fig:multi_plot} quantifies these results for the inferred cloth $C_N(\theta_k)$, texture sliding $C_N'(\theta_k, v)$, and the results of the TSNN $\hat C_N'(\theta_k, v)$.
In Figure \ref{fig:multi_2d}, we repeat these comparisons, except using bilinear interpolation between four camera views.

\subsection{3D Reconstruction} \label{reconstruction_results}
In order to reconstruct the 3D position of a vertex of the ground truth mesh, we take the usual approach of finding rays that pass through that vertex and the camera aperture for a number of cameras.
Then given at least two rays, one can triangulate a 3D point that is minimal distance from all the rays.
We can do this without solving the typical image to image correspondence problem because we know the ground truth texture coordinates for any given vertex.
Thus, we merely have to find the ray that passes through the camera aperture and the ground truth texture coordinate for the vertex under consideration.

\begin{figure}[ht]
\centering
\begin{subfigure}[b]{0.49\linewidth}
    \includegraphics[width=\linewidth]{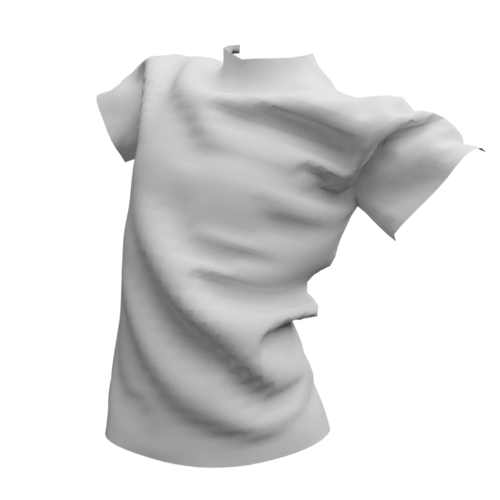}
    \caption{}
    \label{fig:reconstruction_a}
\end{subfigure}
\begin{subfigure}[b]{0.49\linewidth}
    \includegraphics[width=\linewidth]{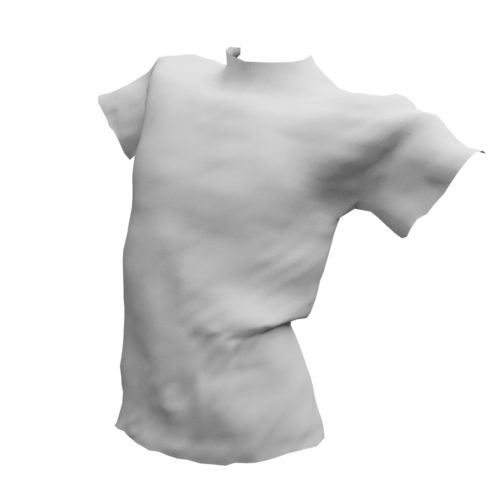}
    \caption{}
    \label{fig:reconstruction_b}
\end{subfigure}
\hfill
\begin{subfigure}[b]{0.49\linewidth}
    \includegraphics[width=\linewidth]{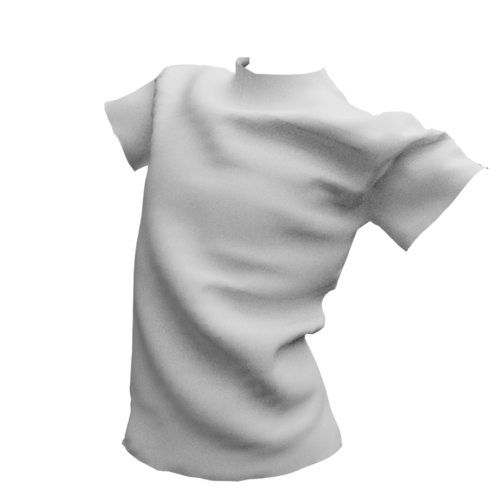}
    \caption{}
    \label{fig:reconstruction_c}
\end{subfigure}
\begin{subfigure}[b]{0.49\linewidth}
    \includegraphics[width=\linewidth]{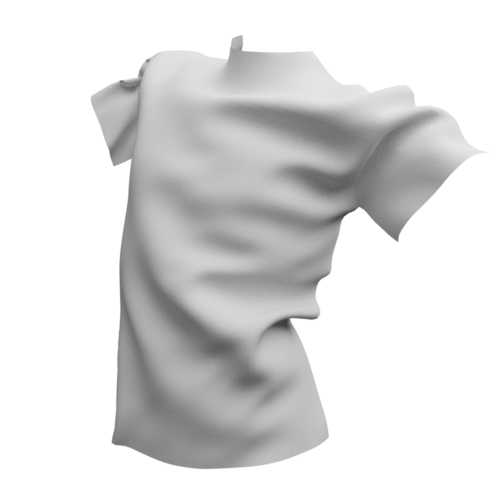}
    \caption{}
    \label{fig:reconstruction_d}
\end{subfigure}
\hfill
\caption{Comparison of the ground truth cloth (a) and inferred cloth (b) to the 3D reconstructions obtained using texture sliding (c) and the TSNN (d). To remove reconstruction noise generated by network inference errors in (d), we used the postprocess from \cite{geng2019coercing}; although, there are many other smoothing options in the literature that one might also consider.}
\label{fig:reconstruction}
\end{figure}

\begin{figure}[ht]
\centering
\begin{subfigure}[b]{0.49\linewidth}
    \includegraphics[width=\linewidth]{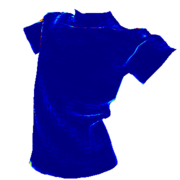}\\
    \vspace{3mm}
    \includegraphics[width=\linewidth]{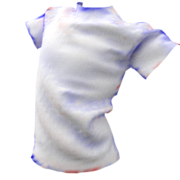}
    \caption{}
    \label{fig:reconstruction_error_a}
\end{subfigure}
\begin{subfigure}[b]{0.49\linewidth}
    \includegraphics[width=\linewidth]{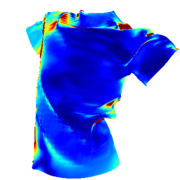}\\
    \vspace{3mm}
    \includegraphics[width=\linewidth]{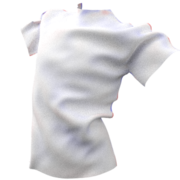}
    \caption{}
    \label{fig:reconstruction_error_b}
\end{subfigure}
\hfill
\caption{Per-pixel errors (top) and local compression/extension energies (bottom) for Figure \ref{fig:reconstruction_c} (a) and Figure \ref{fig:reconstruction_d} (b).}
\label{fig:reconstruction_error}
\end{figure}

To find a ground truth texture coordinate on a texture corrected inferred cloth mesh $C_N'(\theta_k,v)$, or $\hat C_N'(\theta_k,v)$, we first find the triangle containing that texture coordinate.
This can be done quickly by using a hierarchical bounding box structure where the base level boxes around each triangle are defined using the min/max texture coordinates at the three vertices.
Then one can write the barycentric interpolation formula that interpolates the triangle vertex texture coordinates to obtain the given ground truth texture coordinate, and subsequently invert the matrix to solve for the weights.
These weights determine the sub-triangle position of the vertex under consideration (taking care to note that different answers are obtained in 3D space versus screen space, since the camera projection is nonlinear).
Figure \ref{fig:reconstruction} shows the 3D reconstruction of a test set example using texture sliding (Figure \ref{fig:reconstruction_c}) and the TSNN (Figure \ref{fig:reconstruction_d}).
Figure \ref{fig:reconstruction_error} compares the per-pixel errors and local compression/extension energies of Figures \ref{fig:reconstruction_c} and \ref{fig:reconstruction_d}.

\section{Discussion and Future Work}
There are many disparate applications for clothing including for example video games, AR/VR, Hollywood special effects, virtual try-on and shopping, scene acquisition and understanding, and even bullet proof vests and soft armor.
Various scenarios define accuracy or fidelity in vastly different ways.
So while it is typical to state that one cares about more than just the visual appearance (or ``graphics''), often those aiming for predictive capability still make concessions.
For example, wherein \cite{santesteban2019learning} proposes a network that well predicts wrinkles mapped to new body types, the discussion in \cite{lahner2018deepwrinkles} implies that the horizontal wrinkles predicted by \cite{santesteban2019learning} are more characteristic of inaccurate physical simulation than real-world behavior.
Instead, \cite{lahner2018deepwrinkles} strives for more vertical wrinkles to better match their data, but they accomplish this by predicting lighting to match an image while accepting overly smooth geometry. 
And as we have shown in Figure \ref{fig:gt_stress_c}, predicting the correct geometry still allows for rather large errors in the deformation (see \cite{geng2019coercing}).

In light of this, we state the problem of most interest to us: Our aim is to study the efficacy of using deep neural networks to aid in the modeling of material behavior, especially for those materials for which predictive methods do not currently exist because of various unknowns including friction, material parameters (for cloth and body), etc.
Given this goal, we focus on the accurate prediction of material coordinates, which are a super set of deformation, geometry, lighting, visual plausibility, etc.

As demonstrated by the remarkably accurate 3D reconstruction in Figure \ref{fig:reconstruction_c} (see \ref{fig:reconstruction_error_a}), our approach to encoding high frequency wrinkles into lower frequency texture coordinates (\ie texture sliding) works quite well.
It can be used as a post-process to any existing neural network to capture lost details (as long as ground truth and inferred training examples are available); moreover, we showed that trivial subdivision could be used to increase the sampling resolution to limit linearization artifacts. 
The main drawback of our approach is that it relies on triangulation or multi-view stereo in order to construct the final 3D geometry, although this step is not required for AR/VR applications.
This means that one needs to take care when training the texture sliding neural network (TSNN) since inference errors can cause reconstruction noise.
Thus, as future work, we plan on experimenting with the network architecture, the size of the image used in the CNN, the smoothing methods near occlusion boundaries, the amount of subdivision, etc.
In addition, it would be interesting to consider more savvy multiview 3D reconstruction methods (particularly ones that employ DNNs; then, one might train the whole process end-to-end).
Our current solution to addressing multiview reconstruction noise is to simply use the method from \cite{geng2019coercing} as a postprocess to the triangulation of the TSNN results.
As can be seen in Figure \ref{fig:reconstruction_d}, this leads to a high quality reconstruction with many high frequency wrinkles faithful to the ground truth; however, an improved TSNN would lead to more accurate per-pixel texture coordinates than those in Figure \ref{fig:reconstruction_error_b} (top).

\section*{Acknowledgements}
Research supported in part by ONR N00014-13-1-0346, ONR N00014-17-1-2174, and JD.com.
We would like to thank Reza and Behzad at ONR for supporting our efforts into machine learning, as well as Rev Lebaredian and Michael Kass at NVIDIA for graciously loaning us a GeForce RTX 2080Ti to use for running experiments.
We would also like to thank Matthew Cong and Yilin Zhu for their insightful discussions, and Congyue Deng for generating realistic cloth textures.

\section*{Appendix}
\appendix
\section{Dataset Generation} \label{dataset_gen}

\subsection{Topological Considerations} \label{topology}
There are some edge cases that require additional topological consideration.
In particular, the collar, sleeves, and waist are areas where a ray cast to an inferred cloth vertex can intersect with a back-facing triangle on the inside of the ground truth shirt.
We aim to define texture coordinates on inferred cloth vertices so that barycentric interpolation can be used to find the texture coordinates of a ground truth vertex for 3D reconstruction.
However, mixing texture coordinates from the inside and outside of the shirt in a single triangle causes dramatic interpolation error.
In fact, as shown in Figure 2, large errors may occur for any triangle that mixes texture coordinates from geodesically far-away regions.
Thus, we omit such triangles from consideration by omitting a vertex from any edge that connects two geodesically far-away regions.

As a further improvement to our method, one can treat the inside and outside of the shirt as separate meshes, applying texture sliding twice and training two separate networks; moreover, one may take a patch-based approach, applying TS and training a TSNN for each (slightly overlapping) patch of the shirt. 

\subsection{Smoothness Considerations} \label{smoothness}
When training a neural network, more predictable results are obtained when the inferred cloth vertex data is smoother.
Thus, there exists tradeoffs between smoothness and accuracy when assigning texture coordinates.
An edge that connects two geodesically far-away regions introduces a jump discontinuity in the texture coordinates leading to high frequencies in the ground truth data that place increased demands on the network.
Although subdivision adds degrees of freedom along such edges to better sample the high frequency, it is often better to delete such edges entirely by removing one of the edge's vertices.
Recall that any vertex not assigned a ground truth texture coordinate is instead defined via smoothness considerations (see Section \ref{diffusion}) reducing demands on the network.

\begin{figure}[b]
\centering
\begin{subfigure}[b]{\dimexpr0.29\linewidth+20pt\relax}
    \makebox[20pt]{\raisebox{30pt}{\rotatebox[origin=c]{0}{(a)}}}%
    \includegraphics[width=\dimexpr\linewidth-20pt\relax]{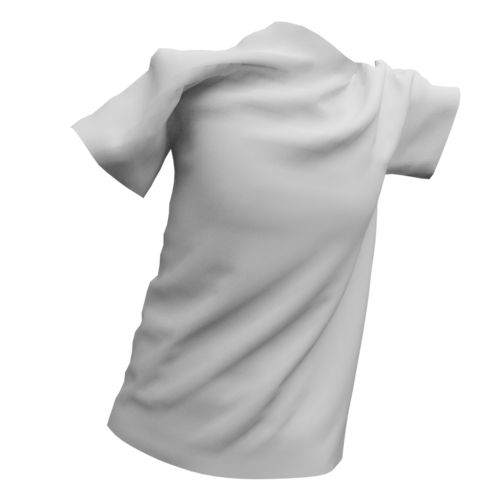}
    \makebox[20pt]{\raisebox{30pt}{\rotatebox[origin=c]{0}{(b)}}}%
    \includegraphics[width=\dimexpr\linewidth-20pt\relax]{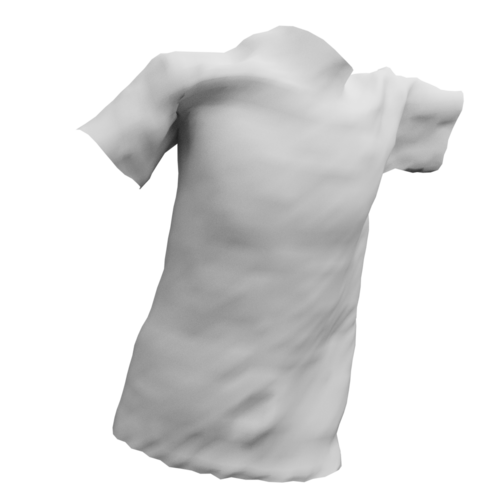}
    \makebox[20pt]{\raisebox{30pt}{\rotatebox[origin=c]{0}{(c)}}}%
    \includegraphics[width=\dimexpr\linewidth-20pt\relax]{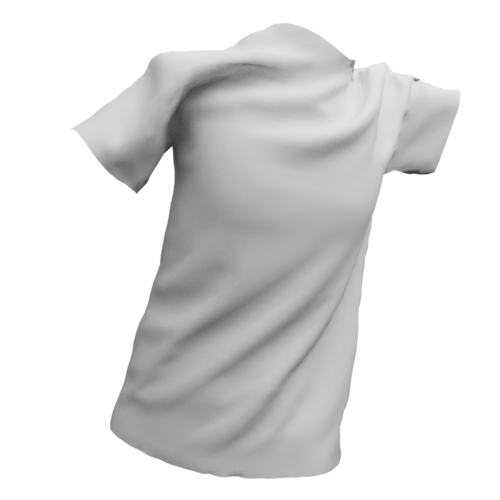}
    \makebox[20pt]{\raisebox{30pt}{\rotatebox[origin=c]{0}{(d)}}}%
    \includegraphics[width=\dimexpr\linewidth-20pt\relax]{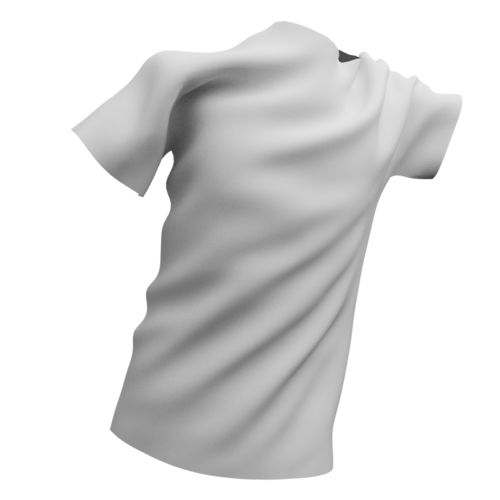}
\end{subfigure}
\begin{subfigure}[b]{0.29\linewidth}
    \includegraphics[width=\linewidth]{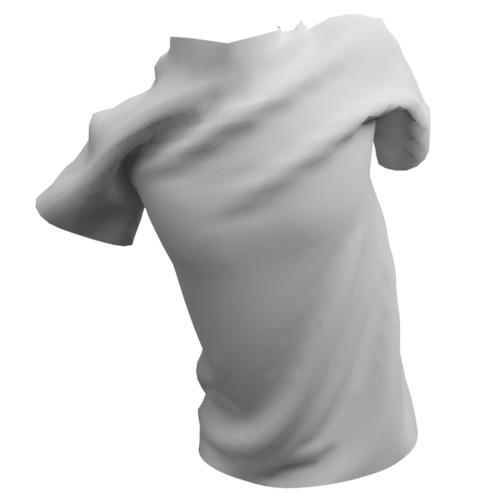}
    \includegraphics[width=\linewidth]{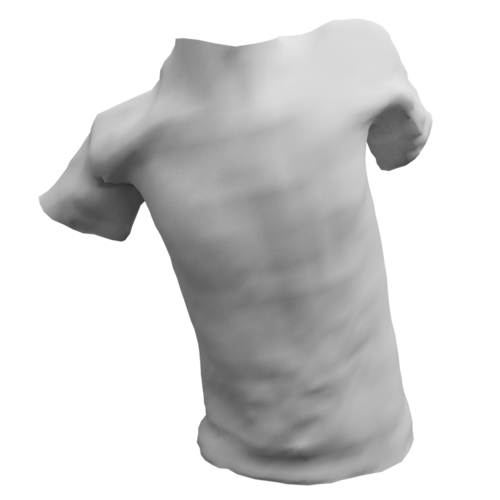}
    \includegraphics[width=\linewidth]{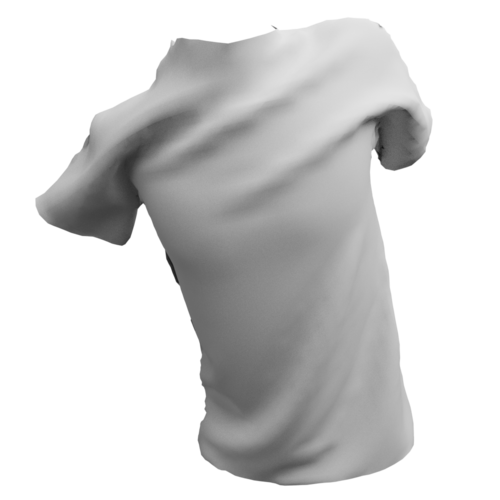}
    \includegraphics[width=\linewidth]{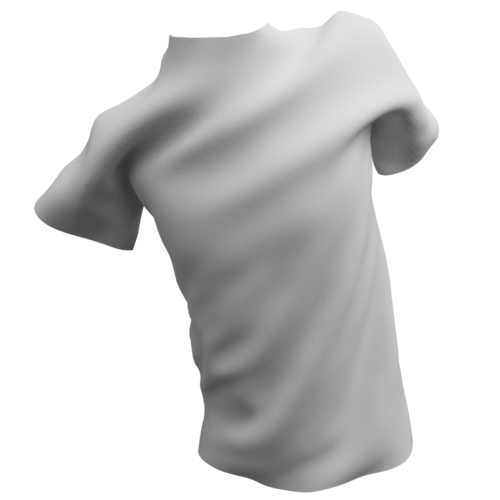}
\end{subfigure}
\begin{subfigure}[b]{0.29\linewidth}
    \includegraphics[width=\linewidth]{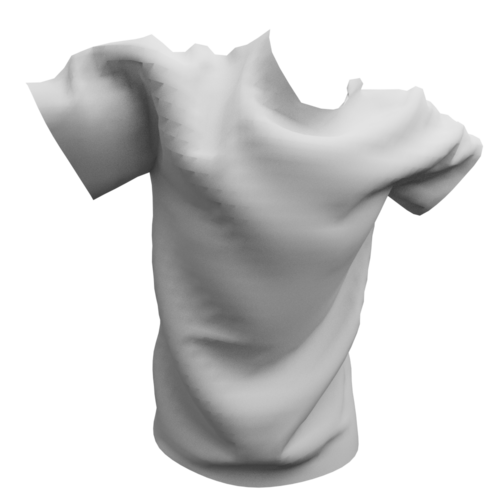}
    \includegraphics[width=\linewidth]{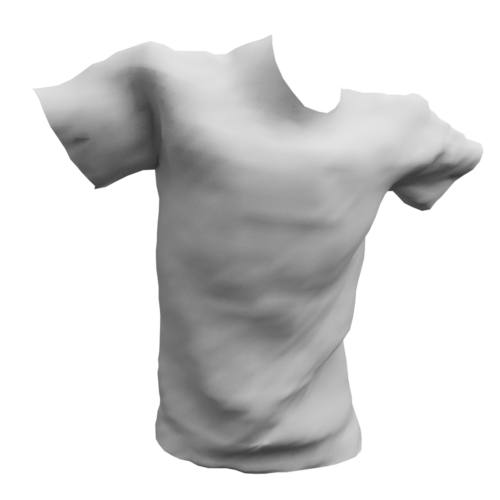}
    \includegraphics[width=\linewidth]{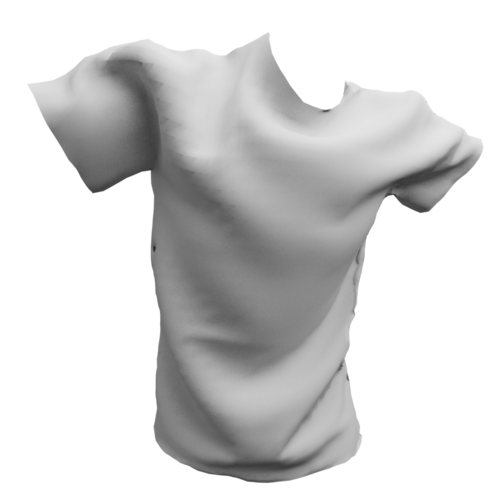}
    \includegraphics[width=\linewidth]{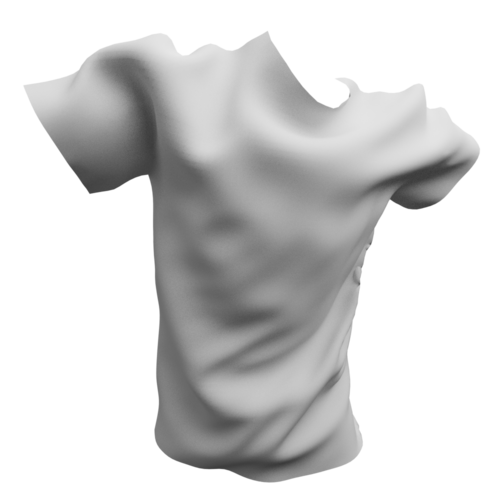}
\end{subfigure}
\hfill
\caption{Comparisons of the ground truth cloth (a) and inferred cloth (b) to the 3D reconstructions obtained using texture sliding (c) and the TSNN (d) for three test set examples. Note that the postprocess in \cite{geng2019coercing} is only applied to (d).}
\label{fig:reconstruction_more}
\end{figure}

\section{3D Reconstruction} \label{reconstruction_extra}
There are a couple of issues with finding the texture coordinates of the ground truth vertices on an inferred cloth mesh whether it be TS or TSNN data.
Firstly, there could be seams in the texture in which case smoothing would be needed near the seam as discussed above in order to avoid degrading the data.
A patch-based approach can be used to alleviate any such seams.
Secondly, seams, smoothing, and non-linearity along the lines of Figure 2 may all contribute to more than one inferred cloth triangle containing the texture coordinates of a ground truth vertex.
This ambiguity can be treated similarly to how correspondence uncertainties are addressed in standard multi-view stereo algorithms.
The straightforward approach is to consider each distinct possibility for each camera in all possible combinations and choose the set of rays that have the least disagreement for triangulation; furthermore, one may also consider the 3D reconstruction of neighboring vertices, material deformation, etc.
Overall, reliance on multi-view stereo does require careful attention when utilizing our method.
As such, we provide a few more examples of 3D reconstruction for examples from the test set in order to demonstrate the efficacy of our approach.
See Figure \ref{fig:reconstruction_more}.

\begin{figure}[t]
\centering
\begin{subfigure}[b]{0.32\linewidth}
    \includegraphics[width=\linewidth]{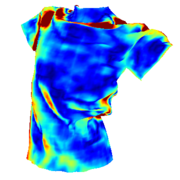}
\end{subfigure}
\begin{subfigure}[b]{0.32\linewidth}
    \includegraphics[width=\linewidth]{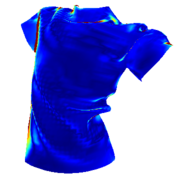}
\end{subfigure}
\begin{subfigure}[b]{0.32\linewidth}
    \includegraphics[width=\linewidth]{figures/sec6/reconstruction/reconstruct_pd_disp_pix_error_16107.png}
\end{subfigure}
\begin{subfigure}[b]{0.32\linewidth}
    \includegraphics[width=\linewidth]{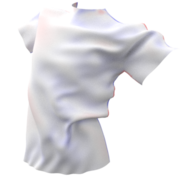}
    \caption{Inferred Cloth}
\end{subfigure}
\begin{subfigure}[b]{0.32\linewidth}
    \includegraphics[width=\linewidth]{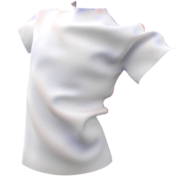}
    \caption{TS}
\end{subfigure}
\begin{subfigure}[b]{0.32\linewidth}
    \includegraphics[width=\linewidth]{figures/sec6/reconstruction/reconstruct_pd_disp_area_16107.png}
    \caption{TSNN}
    \label{fig:reconstruction_fig15}
\end{subfigure}
\hfill
\caption{Comparison of the postprocess from \cite{geng2019coercing} applied to (a) the inferred cloth, (b) the 3D reconstruction from TS data, and (c) the 3D reconstruction from TSNN data. Per-pixel errors (top) and local compression/extension energies (bottom) are shown.}
\label{fig:reconstruction_error_postprocess}
\end{figure}

Instead of applying a standard smoothing algorithm to the somewhat noisy results of the 3D reconstructions of the TSNN data, we used the postprocess from \cite{geng2019coercing}.
This choice was made because of our desire to use neural networks to branch the gap between physical simulations and real-world material behavior.
In order to quantify the impact of the postprocess from \cite{geng2019coercing} on the final results, Figure \ref{fig:reconstruction_error_postprocess} shows the results obtained when applying the postprocess directly to the inferred cloth as compared to applying it to TS and TSNN data.

\section{Novel View Interpolation} \label{multiview_extra}
\begin{figure}[h]
\centering
\includegraphics[width=0.8\linewidth]{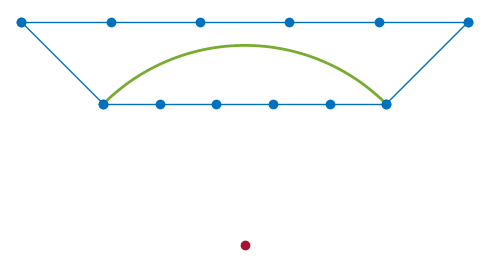}
\caption{Let the red dot represent the center of the cloth mesh. One can interpolate between the 12 cameras (blue dots) on the trapezoid in order to follow the curved camera path.}
\label{fig:interpolate_cameras}
\end{figure}

Interpolating between two cameras, each with TS or TSNN data, has the effect of following a straight-line path.
However, by choosing the camera array and subsequent interpolation carefully one can interpolate along curved paths.
For example in Figure \ref{fig:interpolate_cameras}, one can interpolate between the 12 cameras (represented by blue dots) in order to follow the curved camera path.

\section{Error Analysis (for Patches)}
In this section, we consider each step of the ray intersection algorithm, carefully illustrating the sources of error.
This is done for a single patch consisting of the entire front half of the shirt in order to ensure continuous and unique texture coordinates.
Additionally, this section highlights our patch-based approach, noting that we would utilize this approach on a number of overlapping patches and blend the final results together.
In fact, when considering only a single patch, we modify our nodes from the inferred cloth to only include that patch, ignoring the rest of the vertices and triangles in the mesh.
Similarly, the ground truth cloth is assumed to only consider the data for that patch.
Note that any existing network that predicts cloth vertex positions can be adapted to this patch-based approach as a postprocess applied to their training examples, and that one may readily apply the predicted texture separately to each patch.

\begin{figure}[ht]
\centering
\begin{subfigure}[b]{0.49\linewidth}
    \includegraphics[width=\linewidth]{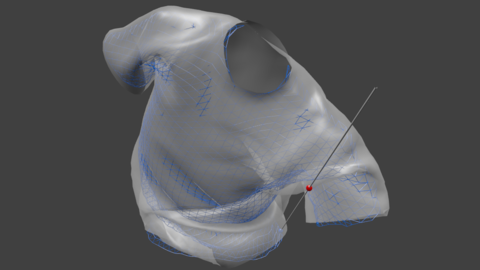}
    \includegraphics[width=\linewidth]{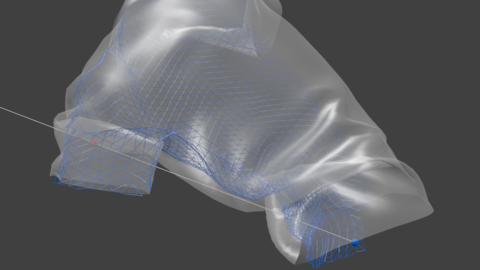}
    \caption{}
    \label{fig:disp_vertex}
\end{subfigure}
\begin{subfigure}[b]{0.49\linewidth}
    \includegraphics[width=\linewidth]{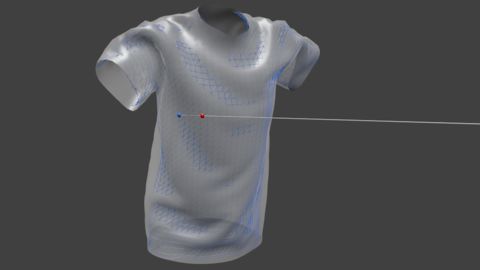}
    \includegraphics[width=\linewidth]{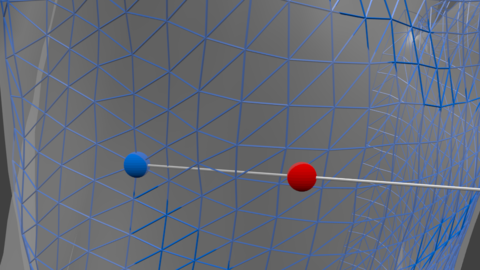}
    \caption{}
    \label{fig:disp_vertex_0.1}
\end{subfigure}
\hfill
\caption{The maximum texture coordinate displacement before (a) and after (b) removing vertices which connect geodesically far away regions. The inferred cloth vertices are drawn in blue, and the ground truth ray intersection points are drawn in red. The wireframe of the inferred cloth is in blue, and the ground truth cloth is in white.}
\label{fig:old_method}
\end{figure}

Along the lines of Section \ref{ray} and Appendix \ref{dataset_gen}, a ray between the camera aperture and each inferred cloth vertex of the patch is intersected with the ground truth cloth, in order to find the ground truth texture coordinates to assign to the inferred cloth vertex.
Recall that the inferred cloth vertex remains unassigned when occluded; however, we modify our definition of occlusion to only consider the inferred cloth patch under consideration.
This allows, for example, one to reconstruct the back half of the shirt with cameras from the front, since the front half of the shirt would not be considered and not be occluded by the back half.
Since we only consider the front half of both the inferred and ground truth cloth, one also does not compute ground truth texture coordinates to be assigned to the inferred vertex when the ray does not intersect the front half patch of the ground truth cloth.
Separating the front and back of the shirt guarantees the inferred cloth patch is assigned texture coordinates from a continuous texture.
This leads to a sub-mesh of assigned texture coordinates $T_N$.
As usual, we remove any edge (by deleting an inferred vertex) which connects geodesically far away regions as indicated by differing texture coordinate values.
See Figure \ref{fig:old_method}.

\subsection{Texture Coordinates}
\begin{figure}[ht]
\centering
\begin{subfigure}[b]{0.49\linewidth}
    \includegraphics[width=\linewidth]{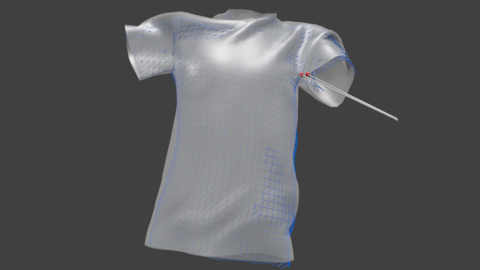}
\end{subfigure}
\begin{subfigure}[b]{0.49\linewidth}
    \includegraphics[width=\linewidth]{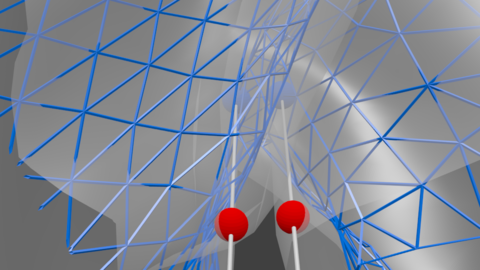}
\end{subfigure}
\hfill
\caption{The edge over the entire training set with the largest change in texture coordinates.}
\label{fig:tex_dist}
\end{figure}

\begin{figure}[ht]
\centering
\begin{subfigure}[b]{0.49\linewidth}
   \includegraphics[width=\linewidth]{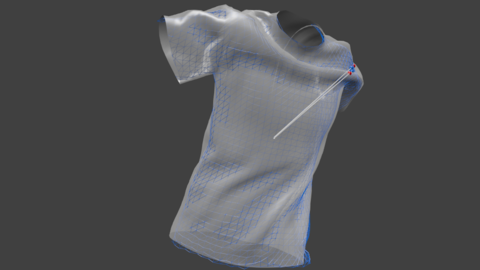}
\end{subfigure}
\begin{subfigure}[b]{0.49\linewidth}
    \includegraphics[width=\linewidth]{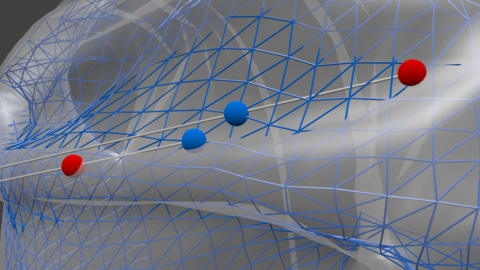}
\end{subfigure}
\hfill
\caption{The edge over the entire training set with the largest ground truth intersection Euclidean distance.}
\label{fig:euc_dist}
\end{figure}

To quantify the worst case texture sliding scenarios, we first consider $T_N$ for every pose $\theta_k$ and camera view $v_p$ used in training.
The edge with the largest difference in texture coordinates (as a proxy for geodesic distances) is shown in Figure \ref{fig:tex_dist}.
We do the same for Euclidean distances along every edge to connected vertices in Figure \ref{fig:euc_dist}.


\subsection{Texture Coordinate Displacements}
\begin{figure}[ht]
\centering
\begin{subfigure}[b]{0.49\linewidth}
    \includegraphics[width=\linewidth]{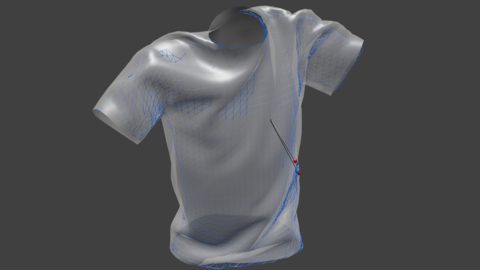}
\end{subfigure}
\begin{subfigure}[b]{0.49\linewidth}
    \includegraphics[width=\linewidth]{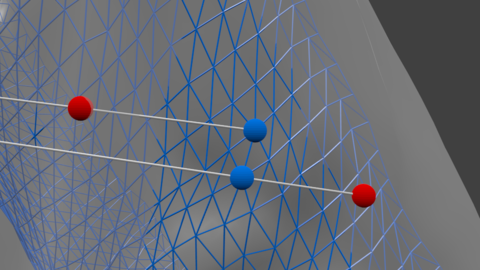}
\end{subfigure}
\hfill
\caption{The edge over the entire training set with the largest change in texture coordinate displacements.}
\label{fig:disp_edge}
\end{figure}

In order to fill in unassigned vertices for the patch under consideration, we show the most extreme behavior of texture sliding over all $(\theta_k, v_p)$.
First, we compute the maximal value of $||d_{v_p}(\theta_k)||$ among all $(\theta_k, v_p)$ pairs, \ie where maximal texture sliding occurs in our training set.
See Figure \ref{fig:disp_vertex_0.1}.
We also compute $\Delta d_{v_p}(\theta_k)$ along each assigned edge in order to ascertain the biggest jump (indicating high frequency) that would be seen by the TSNN.
The edge with the maximal $||\Delta d_{v_p}(\theta_k)||$ over all $(\theta_k, v_p)$ pairs is shown in Figure \ref{fig:disp_edge}.
Note that one may obtain better smoothness when training the TSNN by not assigning vertices where $d_{v_p}(\theta_k)$ is too large or one of the vertices of an edge where $\Delta d_{v_p}(\theta_k)$ is too large.

\subsection{Smoothness Considerations}
As long as extrapolation is done smoothly to assign texture coordinates to the remaining vertices, there should be no new extrema in $\Delta T_N(\theta_k,v_p)$ and $\Delta d_{v_p}(\theta_k)$.
After applying smoothing, we verify that the largest $\Delta T_N(\theta_k,v_p)$ and $\Delta d_{v_p}(\theta_k)$ are the same as before.

\section{TSNN -- Additional Experiments} \label{tsnn_extra}
Table \ref{tab:net_comp} shows additional TSNN results after applying a displacement threshold to the TS dataset.
In addition, in Table \ref{tab:net_comp_breakdown} we decompose the TSNN errors based on whether vertices were assigned via our ray intersection method or extrapolation.
Results indicate that training separate networks for smooth and wrinkled regions of the cloth may be a promising avenue for future work.

\begin{table}[ht]
\begin{center}
\begin{tabular}{|c|c|}
\hline
Network & SqrtMSE ($\times 10^{-3}$)\\
\hline\hline
TSNN & 15.058 $\pm$ 6.5256 \\
TSNN + subdivision & 14.926 $\pm$ 6.5918 \\
\hline
\end{tabular}
\end{center}
\caption{Comparisons of per-pixel SqrtMSE for the test set after applying a threshold to the ground truth TS displacements.}
\label{tab:net_comp}
\end{table}

\begin{table}[ht]
\begin{center}
\begin{tabular}{|c|c|c|}
\hline
 & TSNN (original) & TSNN (threshold) \\
\hline\hline
Ray Intersection & 11.670 $\pm$ 3.2160 & 10.958 $\pm$ 2.9056 \\
Extrapolation & 39.564 $\pm$ 27.087 & 95.200 $\pm$ 48.406 \\
Combination & 14.279 $\pm$ 4.5970 & 15.405 $\pm$ 5.5923 \\
\hline
\end{tabular}
\end{center}
\caption{Breakdown of the TSNN errors ($\times 10^{-3}$) in Tables \ref{tab:net_mse} and \ref{tab:net_comp} based on whether each pixel contains vertices assigned via ray intersection, diffusion, or a combination of both.}
\label{tab:net_comp_breakdown}
\end{table}

\clearpage

{\small
\bibliographystyle{ieee}
\bibliography{egbib}
}

\end{document}